%% file: main.tex
\documentclass{article}

\usepackage[preprint,nonatbib]{techreport}

\usepackage[utf8]{inputenc} 
\usepackage[T1]{fontenc}    
\usepackage{hyperref}       
\usepackage{url}            
\usepackage{booktabs}       
\usepackage{amsfonts}       
\usepackage{nicefrac}       
\usepackage{microtype}      
\usepackage{xcolor}         
\usepackage[table]{xcolor}
\usepackage{ulem}

\usepackage{amsmath}
\usepackage{xspace}
\usepackage{graphicx}
\usepackage{algorithm}
\usepackage{algorithmic}
\usepackage{multirow}
\usepackage{bbding}
\usepackage{caption}
\usepackage{tabularx}
\usepackage{wrapfig}
\usepackage[most]{tcolorbox}
\usepackage{sectsty}
\usepackage{xcolor}
\usepackage[most]{tcolorbox}

\newcommand{\method}{FashionChameleon\xspace}

\definecolor{taobaocolor}{RGB}{220, 70, 0} 
\definecolor{xmucolor}{RGB}{68, 114, 196} 

\sectionfont{\color{taobaocolor}}
\subsectionfont{\color{taobaocolor}}
\subsubsectionfont{\color{taobaocolor}}

\title{\includegraphics[height=1.0em]{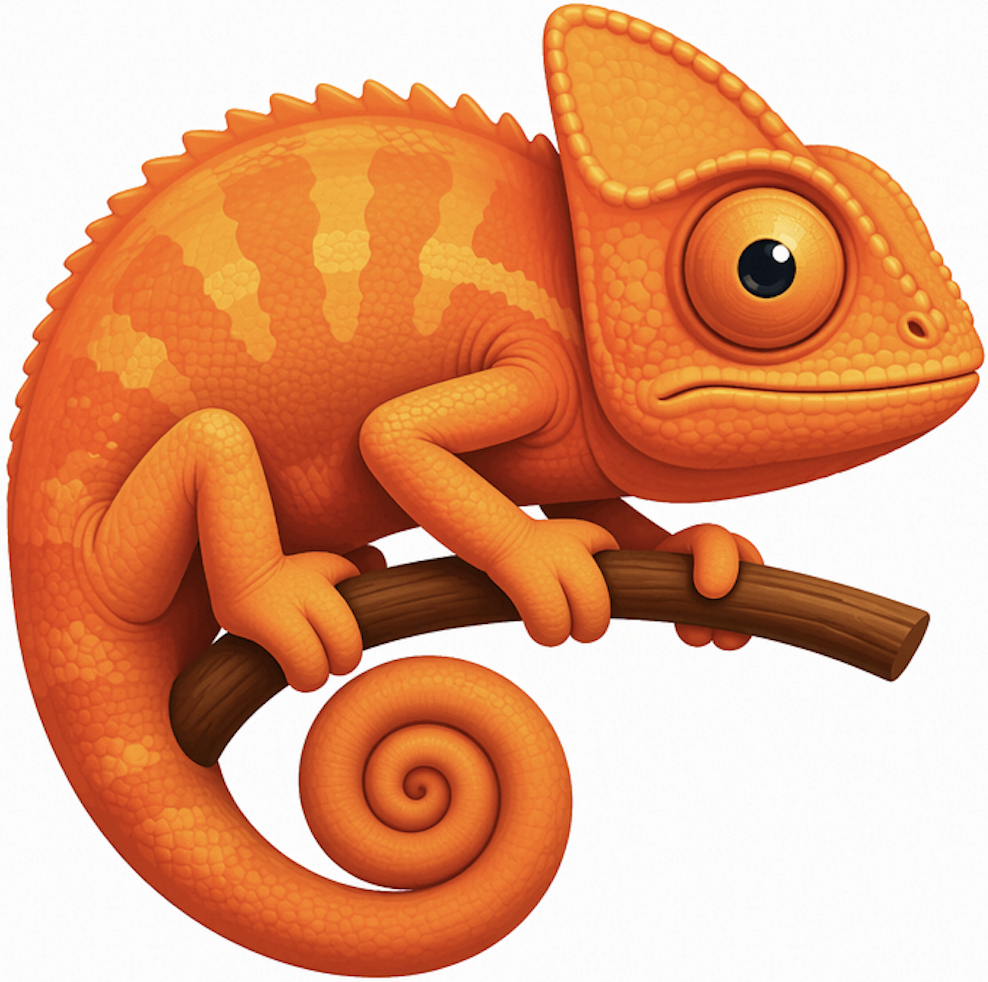} \method: Towards Real-Time and Interactive Human-Garment Video Customization}

\author{
  \textbf{Quanjian Song\textsuperscript{\rm 1, 2}, \; Yefeng Shen\textsuperscript{\rm 2}, \; Mengting Chen\textsuperscript{\rm 2,$*$}, \; Hao Sun\textsuperscript{\rm 2},} \\ \textbf{Jinsong Lan\textsuperscript{\rm 2}, \; Xiaoyong Zhu\textsuperscript{\rm 2}, \; Bo Zheng\textsuperscript{\rm 2}, \; Liujuan Cao\textsuperscript{\rm 1}} \vspace{0.03cm} \\
  \textsuperscript{\rm 1}Xiamen University \quad \textsuperscript{\rm 2}Alibaba Group \vspace{0.07cm} \\
  Project Page: \href{https://quanjiansong.github.io/projects/FashionChameleon/}{\includegraphics[width=0.8em]{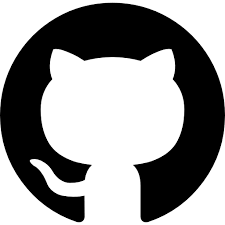}}
}

\begin{document}

\maketitle

\renewcommand{\thefootnote}{\fnsymbol{footnote}}
\footnotetext[1]{Project leader.}

\input{sections/0-abs}

\begin{center}
  \includegraphics[width=\textwidth]{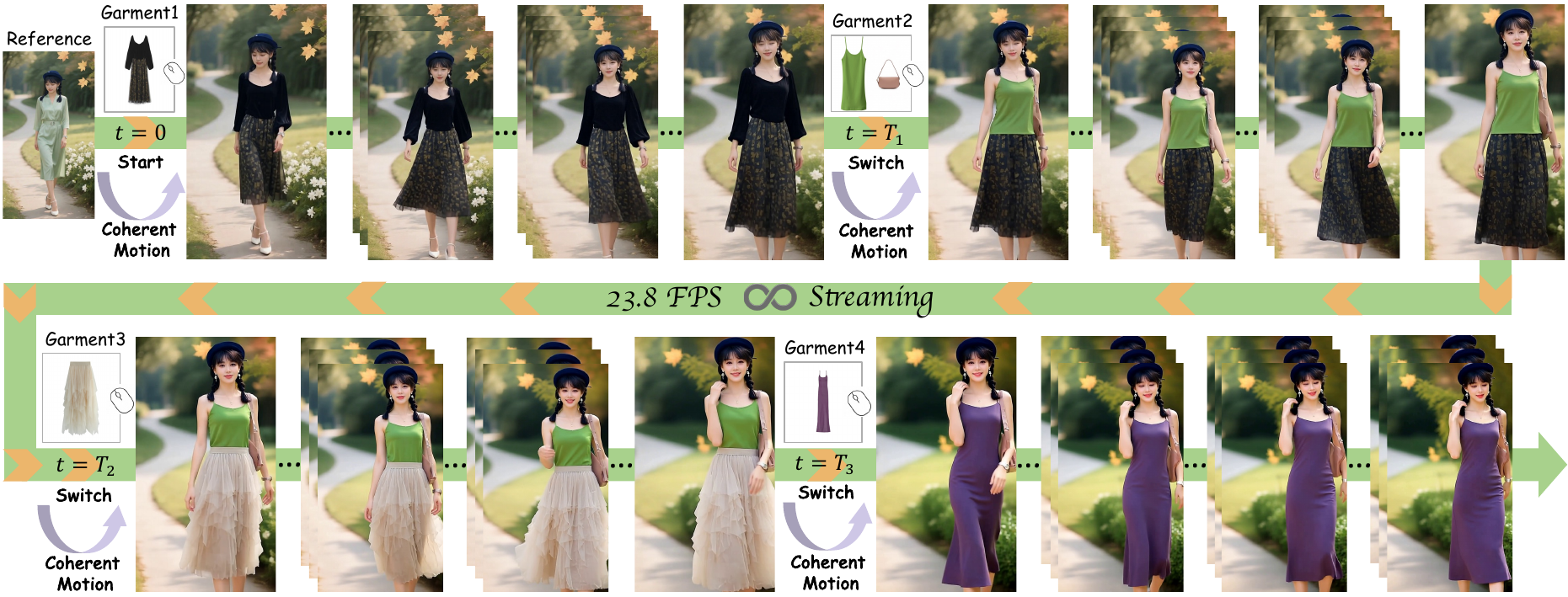}
  \captionof{figure}{
  Given a reference image and a sequence of garment images, \method generate customized videos in a streaming and interactive manner, where users can interactively switch garments during generation while preserving coherent motion, achieving $23.8$ FPS real-time generation.
}
  \label{fig:teaser}
\end{center}

\input{sections/1-intro}
\input{sections/2-related_works}
\input{sections/3-method}
\input{sections/4-exp}
\input{sections/5-conclusion}

\bibliographystyle{IEEEtran}
\bibliography{reference}


\clearpage
\input{sections/X-suppl}


\end{document}

%% file: sections/0-abs.tex
\newenvironment{customabstract}{
    \begin{tcolorbox}[
        enhanced,
        colframe=taobaocolor,
        colback=white,
        boxrule=0.7pt,
        arc=12pt,
        auto outer arc,
        left=15pt,
        right=15pt,
        top=10pt,
        bottom=15pt,
        breakable
    ]
    \begin{center}
        \Large\bfseries\color{taobaocolor} Abstract
    \end{center}
    \vspace{0.2em}
}{
    \par\vspace{1em}
    \hfill \textit{Date: May 18, 2026}
    \end{tcolorbox}
}

\begin{customabstract}
Human-centric video customization, particularly at the garment level, has shown significant commercial value. However, existing approaches cannot support low-latency and interactive garment control, which is crucial for applications such as e-commerce and content creation.
This paper studies how to achieve interactive multi-garment video customization while preserving motion coherence using only single-garment video data.
We present \textbf{\method}, a real-time and interactive framework for human-garment customization in autoregressive video generation, where users can interactively switch garment during generation.
\method consists of three key techniques:
(i) Instead of training on multi-garment video data, we train a \textbf{Teacher Model with In-Context Learning} on a single reference–garment pair. By retaining the \textit{image-to-video training paradigm} while \textit{enforcing a mismatch between the reference and garment image}, the model is encouraged to implicitly preserve coherence during single-garment switching.
(ii) To achieve consistency and efficiency during generation, we introduce \textbf{Streaming Distillation with In-Context Learning}, which fine-tunes the model with \textit{in-context teacher forcing} and improves extrapolation consistency via \textit{gradient-reweighted distribution matching distillation}.
(iii) To extend the model for interactive multi-garment video customization, we propose \textbf{Training-Free KV Cache Rescheduling}, which includes \textit{garment KV refresh}, \textit{historical KV withdraw}, and \textit{reference KV disentangle} to achieve garment switching while preserving motion coherence.
Our \method uniquely supports interactive customization and consistent long-video extrapolation, while achieving real-time generation at 23.8 FPS on a single GPU, 30-180$\times$ faster than existing baselines.
\end{customabstract}

%% file: sections/1-intro.tex
\section{Introduction}
Driven by advances in diffusion models~\cite{ho2020denoising, lipman2022flow}, text-to-video and image-to-video generation~\cite{yang2024cogvideox, kong2024hunyuanvideo, wan2025wan} have become prominent directions.
However, these approaches condition only on a simple prompt or an initial frame, which limits their applicability in real-world scenarios~\cite{li2025realcam,fu2024drivegenvlm,song2025worldwander}.
To overcome this limitation, recent work has explored various customized video generation, in which visual concepts are injected into the generation process through user-provided reference images.
One representative setting is subject-to-video (S2V)~\cite{wang2026customvideo,chen2024disenstudio,he2024id,yuan2025identity,liu2025phantom,vace,xue2025stand}  customization, which aims to ensure that subjects in generated videos remain consistent with the given reference images.
With the advances of Diffusion Transformers (DiT)~\cite{peebles2023scalable,yang2024cogvideox,kong2024hunyuanvideo,wan2025wan}, subsequent works~\cite{li2025bindweave,deng2025magref,fei2025skyreels,zhang2025kaleido} extend S2V customization to multi-reference settings, enabling more flexible control in complex scenes.

\begin{wrapfigure}{r}{0.48\columnwidth}
    \centering
    \includegraphics[width=\linewidth]{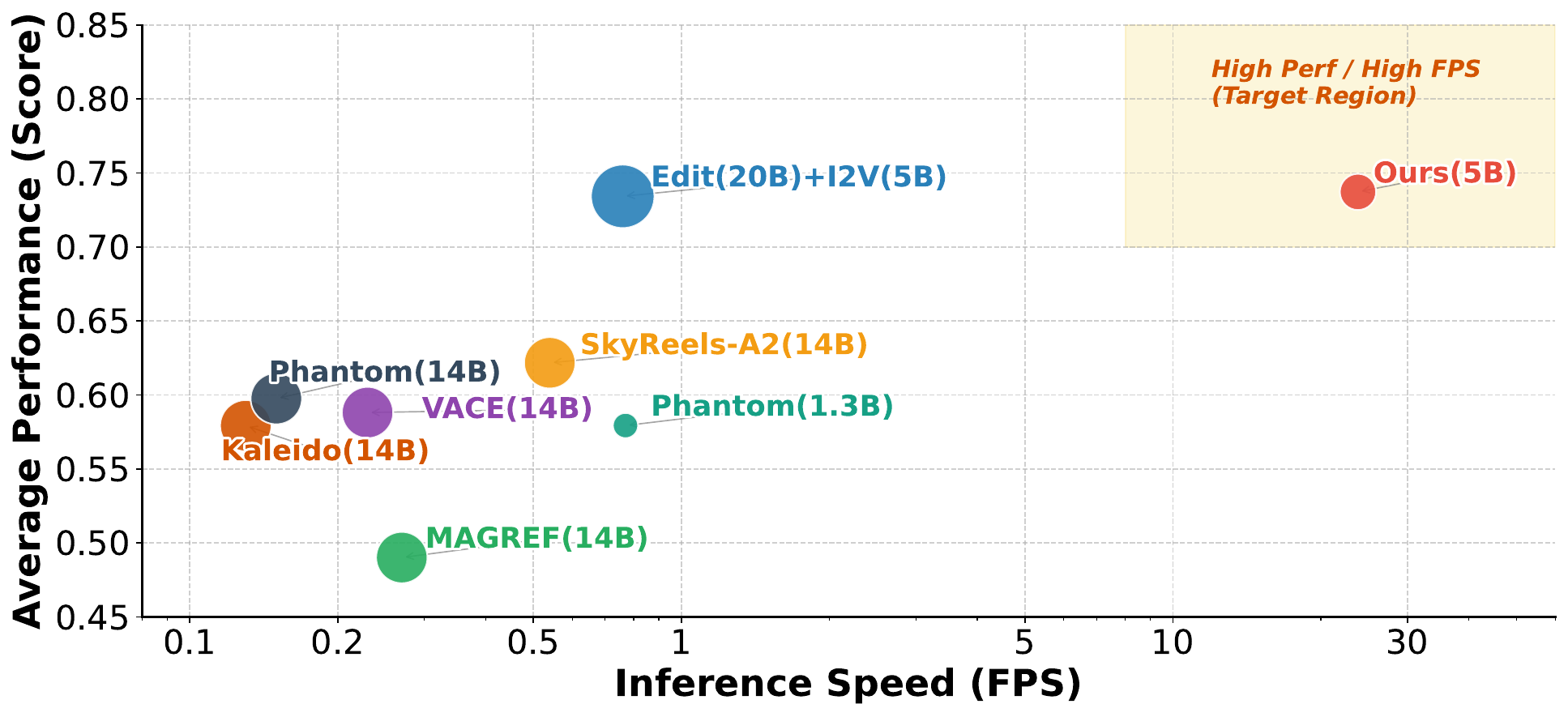}
    \caption{
    Average performance (Cur., GME, Amp., Smoo., and VQ) and inference speed comparison across different approaches.
    }
    \label{fig:intro}
    \vspace{-0.7em}
\end{wrapfigure}

Despite this progress, existing customization methods mainly focus on human-centric subject consistency, with comparatively less emphasis on fine-grained human attributes.
Among these attributes, garment-level customization is particularly desirable in practical applications such as filmmaking~\cite{wang2024motionctrl,song2025lightmotion}, e-commerce~\cite{lari2022artifical} and entertainment~\cite{li2023photomaker,song2025univst,zhang2025objectadd}, where users often require low-latency, streaming, and interactive control over garments.
Given the recent success of hybrid autoregressive generation~\cite{yin2025slow,huang2025self,zhu2026causal} in diverse domains~\cite{zhuang2025flashvsr,huang2025live,shin2025motionstream}, we are inspired to ask: \textit{Can this paradigm be extended to the customization domain?}
In this work, we formulate \textbf{streaming and interactive human-garment video customization} and pinpoint three key challenges:
(i) \textit{Single-to-multiple generalization.}
Video data with multi-garment switching are typically difficult to obtain. How to effectively exploit single-garment data for interactive multi-garment video customization remains a significant challenge.
(ii) \textit{Consistency and efficiency.}
Although distillation from bidirectional to autoregressive generation improves inference efficiency, it also introduces error accumulation during self-rollout.
In human-centric scenarios, it is important to maintain identity and motion consistency while achieving efficiency during streaming generation.
(iii) \textit{Coherent interaction.}
Interactive video customization requires dynamically switching a character's garments during generation.
Ensuring seamless garment transitions while preserving continuous human motion remains challenging.

In this paper, we introduce \textbf{\method}, a real-time and interactive framework that enables human-garment customization in autoregressive video generation (see Figure~\ref{fig:teaser}), where users can interactively switch garments during generation while maintaining coherent human motion.

(i) Rather than directly training a teacher model on multi-garment video data, we train a \textbf{Teacher Model with In-Context Learning} to process a reference image paired with a garment image.
Notably, we retain the \textit{image-to-video training paradigm} while ensuring that \textit{the garment worn by the reference person differs from the target garment}.
This enables the model to implicitly preserve coherence during single-garment switching, laying the foundation for interactive multi-garment switching.

(ii) To achieve consistency and efficiency during streaming video generation, we introduce \textbf{Streaming Distillation with In-Context Learning}.
Specifically, it fine-tunes the model with \textit{in-context teacher forcing} to eliminate the data-intensive ODE initialization, and incorporates \textit{gradient-reweighted distribution matching distillation} to improve consistency in long-video extrapolation.

(iii) To extend the model for interactive multi-garment video customization, we propose \textbf{Training-Free KV Cache Rescheduling}.
Specifically, it first perform \textit{garment KV refresh} to switch garments during inference, then apply \textit{historical KV withdraw} to suppress outdated garment in historical frames, and utilize \textit{reference KV disentangle} to preserve coherent human motion during garment-switching.

To further support teacher model pre-training and streaming distillation post-training, we propose a high-quality data curation pipeline with four stages: \textit{general coarse-to-fine video filtering}, \textit{static-dynamic video captioning}, \textit{fine-grained garment image extraction}, and \textit{adaptive reference image extraction}.
Qualitative and quantitative experiments on the proposed HGC-Bench show that our \method\ is superior to existing baselines while achieving real-time 720p customization at 23.8 FPS on a single H200 GPU (see Figure~\ref{fig:intro}). 
Additional experiments on interactive multi-garment video customization and consistent long-video extrapolation further highlight its unique capabilities.

%% file: sections/2-related_works.tex
\section{Related Works}

\noindent
\textbf{Subject-to-Video Customization.}
Subject-to-Video (S2V) aims to preserve subjects specified by reference images for customized video generation.
Early approaches~\cite{wang2026customvideo,chen2024disenstudio} rely on few-shot tuning, while later works~\cite{he2024id,yuan2025identity} improve generalization by fine-tuning U-Net-based models.
With the rise of diffusion transformers (DiT)~\cite{peebles2023scalable,bao2023all}, subsequent methods~\cite{vace,fei2025skyreels,xue2025stand,liu2025phantom} focus on human-centric customization, with improved identity preservation, editing flexibility, and text-image alignment.
Recent works extend this paradigm to multi-reference customization:
MAGREF~\cite{deng2025magref} supports any-reference generation via subject disentanglement, while BindWeave~\cite{li2025bindweave} and Kaleido~\cite{zhang2025kaleido} improve multi-entity grounding and reference integration in complex scenes.
Despite this progress, they suffer from high inference latency and limited interactivity, which are crucial for practical user experience.
In contrast, our \method achieves real-time and interactive customization.

\noindent
\textbf{Hybrid Autoregressive Video Generation.}
Recent hybrid autoregressive video generation methods~\cite{chen2024diffusion,yin2025slow,huang2025self,zhu2026causal} combine diffusion-based frame modeling~\cite{kong2024hunyuanvideo,yang2024cogvideox,wan2025wan} with autoregressive prediction across frames~\cite{kondratyuk2023videopoet,sun2024autoregressive}, balancing fidelity and efficiency.
CausVid~\cite{yin2025slow} leverages distribution matching distillation (DMD)~\cite{yin2024improved} to distill a slow bidirectional teacher into a few-step autoregressive student, avoiding training from scratch.
Furthermore, Self Forcing~\cite{huang2025self} conditions the model on its own rolled-out frames instead of ground-truth frames, thereby fundamentally solving the training-inference mismatch.
Building on this paradigm, Rolling Forcing~\cite{liu2025rolling} accelerates inference, Reward Forcing~\cite{lu2025reward} improves motion dynamics, Infinity-RoPE~\cite{yesiltepe2025infinity} enables stable long-video generation, and Causal Forcing~\cite{zhu2026causal} reduces distribution mismatch during ODE initialization.

\noindent
\textbf{Applications of Streaming Video Generation.}
Benefiting from low latency and interactive inference, hybrid autoregressive generation has been adopted in various downstream tasks.
LiveAvatar~\cite{huang2025live}, FlashVSR~\cite{zhuang2025flashvsr}, MotionStream~\cite{shin2025motionstream}, and LongLive~\cite{yang2025longlive} extend this paradigm to audio-driven avatar generation, video super-resolution, interactive motion-controlled generation, and interactive prompt-controlled generation, respectively.
More recently, popular video world models, such as Vid2World~\cite{huang2025vid2world}, Yume~\cite{mao2025yume}, WorldPlay~\cite{sun2025worldplay}, and Matrix-Game~\cite{zhang2025matrix} further exploit it for interactive virtual worlds.
However, these works mainly consider continuous control signals such as audio, motion, or mouse/keyboard inputs. To the best of our knowledge, no research has yet explored streaming applications in customized video generation tasks, particularly those involving discrete control signals like garment manipulation. Our work seeks to address this gap.

\section{Preliminary}

\noindent
\textbf{Video Diffusion Models.}
The advanced video diffusion generation typically consists of a variational encoder–decoder pair $\langle \mathcal{E}, \mathcal{D} \rangle$ along with a transformer-based predict network $v_{\theta}$.
During training, the encoder $\mathcal{E}$ transforms a video with $F$ frames into a latent sequence $\mathbf{z}_{0}^{1:f}$ with $f$ frames, where $f = \frac{F - 1}{4} + 1$.
According to flow matching~\cite{lipman2022flow}, the forward process is defined as a linear interpolation between the data distribution and a standard normal distribution, as follows:
\begin{equation}
    z_t^{1:f} = (1 - t) \cdot z_0^{1:f} + t \cdot \epsilon^{1:f},
\label{eq:1}
\end{equation}
where $t$ is a random timestep and $\epsilon^{1:f} \sim \mathcal{N}(0, I)$.
For the noisy latent $z_t^{1:f}$, we utilize the predict network $v_{\theta}$ to regress the conditional vector field via conditional flow matching~\cite{lipman2022flow} loss:
\begin{equation}
    \min_{\theta} \mathbb{E}_{t \sim \mathcal{U}(0,1)} \| v_{\theta}(z_{t}^{1:f}, t, c) - v \|_2^2,
\label{eq:2}
\end{equation}
where $v = \epsilon^{1:f} - z_0^{1:f}$ denotes the target vector field, and $c$ represents the conditional signals.

\noindent
\textbf{Hybrid Autoregressive Video Generation.}
Given a video with $F$ frames $\mathcal{V}^{1:F} = \langle \mathcal{V}^1, \mathcal{V}^2, \ldots, \mathcal{V}^F \rangle$, CausVid~\cite{yin2025slow} proposes to factorizes the joint distribution as $p(\mathcal{V}^{1:F}) = \prod_{i=1}^{F} p(\mathcal{V}^{i} \mid \mathcal{V}^{<i})$, where each conditional distribution $p(\mathcal{V}^{i} \mid \mathcal{V}^{<i})$ is modeled by the diffusion models where each frame/chunk is generated autoregressively.
Self-Forcing~\cite{huang2025self} further improves this paradigm with self-rolling, conditioning on self-generated rather than ground-truth history to better align training with inference.
To avoid training from scratch, most methods distill multi-step bidirectional teacher models into few-step autoregressive student models via Distribution Matching Distillation (DMD)~\cite{yin2024improved}.
Specifically, DMD minimizes an approximate KL divergence between the student distribution estimated by $s_{\text{fake}}$ and the data distribution estimated by $s_{\text{real}}$. This process can be formulated as follows:
\begin{equation}
\nabla \mathcal{L}_{\text{DMD}} = - \mathbb{E}_{t} \Biggl[ 
    \int \Bigl( 
        s_{\text{real}}(\phi(G(\epsilon), t), t) - s_{\text{fake}}(\phi(G(\epsilon), t), t) 
    \Bigr) \cdot \frac{d G_{\theta}(\epsilon)}{d\theta} \, d\epsilon
\Biggr],
\label{eq:3}
\end{equation}
where $\epsilon \sim  \mathcal{N}(0, I)$, $G_{\theta}$ denotes student model, and $\phi(\cdot,t)$ represents forward diffusion at timestep $t$ defined in Eq.\,\ref{eq:1}.
During distillation, $G_{\theta}$ and $s_{\text{fake}}$ are updated while $s_{\text{real}}$ remains frozen.

%% file: sections/3-method.tex
\section{Methodology}
\begin{figure}[t]
    \centering
    \includegraphics[width=0.98\linewidth]{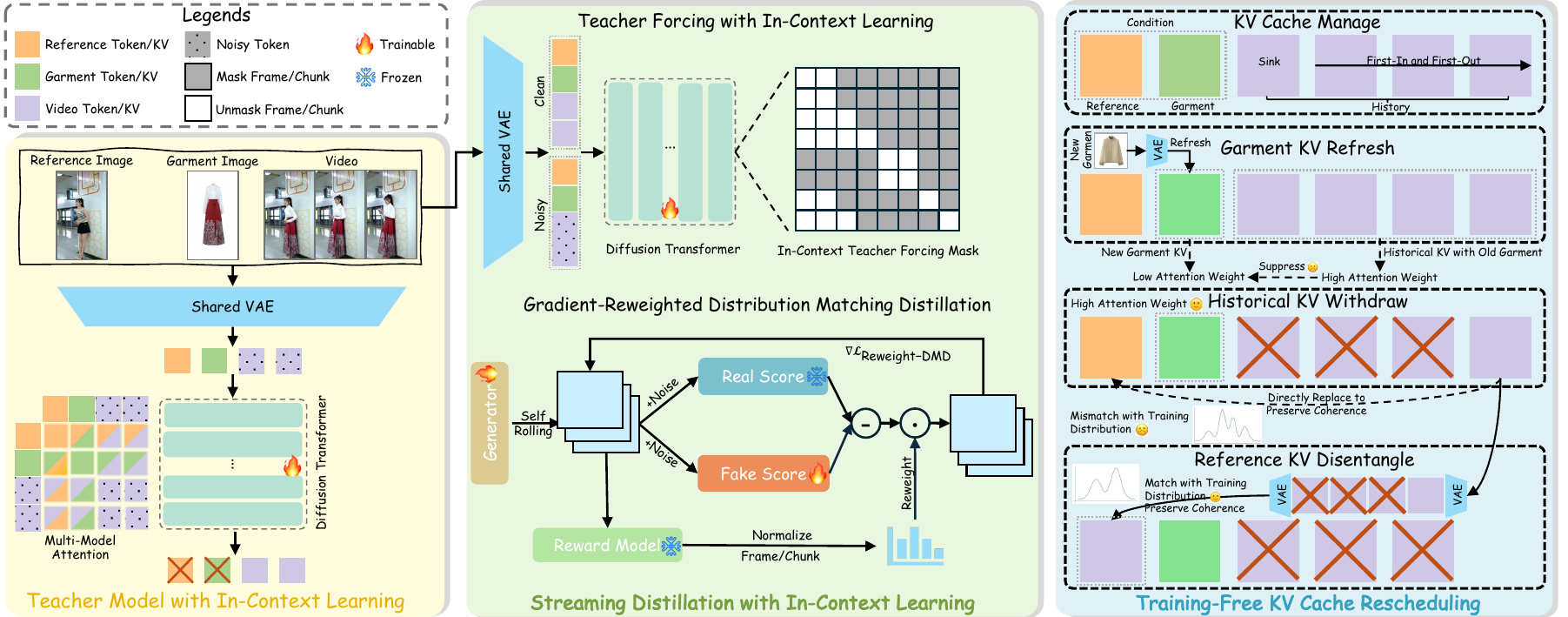}
    \caption{
    Overall pipeline of \method: \textit{Teacher Model with In-Context Learning}, \textit{Streaming Distillation with In-Context Learning}, and \textit{Training-Free KV Cache Rescheduling}.
    }
    \label{fig:overall_pipeline}
    \vspace{-1.0em}
\end{figure}

In this work, we propose \textbf{\method}, a real-time and interactive framework that enables human-garment customization in autoregressive video generation.
Given a reference image $I^{\text{src}}$ and a sequence of $N$ garment images $\langle I^{\text{gar}_1}, \ldots, I^{\textit{gar}_N} \rangle$, our goal is to generate videos in a streaming manner, where each garment is applied to the character at different moments while ensuring coherent human motion.
In Sec.\,\ref{sec:3-1}, we first train a \textbf{Teacher Model with In-Context Learning} conditioned on a reference image and a single garment image.
In Sec.\,\ref{sec:3-2}, we introduce \textbf{Streaming Distillation with In-Context Learning}, featuring an \textit{in-context teacher forcing mask} technique for stable training and a \textit{gradient-reweighted distribution matching distillation} strategy to improve extrapolation consistency.
In Sec.\,~\ref{sec:3-3}, we propose \textbf{Training-Free KV Cache Rescheduling}, which consists of \textit{garment KV refresh}, \textit{historical KV withdraw}, and \textit{reference KV disentangle}, enabling seamless garment switching while maintaining motion coherence.
In Sec.\,\ref{sec:3-4}, we develop a \textbf{High-Quality Data Curation Pipeline} to further support training.
The overall pipeline of \method is shown in Figure~\ref{fig:overall_pipeline}.

\subsection{Teacher Model with In-Context Learning}
\label{sec:3-1}
To enable real-time and interactive human-garment video customization, we first train a bidirectional teacher model conditioned on a reference image and a single garment image.
Unlike prior works~\cite{huang2025live,zhuang2025flashvsr,shin2025motionstream} that rely on auxiliary encoders to process continuous signals, we adopt in-context learning within a unified backbone network to process discrete reference and garment images, eliminating the auxiliary encoders.
\textbf{Notably}, we retain the \textit{image-to-video (I2V) training property}, such that the first generated frame stays consistent with the reference frame, except for the garment information.
\textbf{Meanwhile}, we ensure that \textit{the garment worn by the reference person differs from the target garment}.
This implicitly enables the model to learn single-garment switching while maintaining coherence.

\noindent
\textbf{Shared Latent Space with Varying Noise Levels.}
During training process, a given video $\mathcal{V}$ is encoded into a latent representation $z_{0}^{v}$ by the VAE encoder $\mathcal{E}$.
Instead of introducing an additional encoder, we reuse $\mathcal{E}$ to separately encode the reference image $I^{\text{src}}$ and the garment image $I^{\text{gar}}$ into latent representations $z^{\text{src}}_{0}$ and $z^{\text{gar}}_{0}$.
The whole process can be formulated as follows:
\begin{equation}
    z^{v}_{0} = \mathcal{E}(\mathcal{V});\quad
    z^{\text{src}}_{0} = \mathcal{E}(I^{\text{src}});\quad
    z^{\text{gar}}_{0} = \mathcal{E}(I^{\text{gar}}).
\end{equation}
In this way, all latents can share semantic space without introducing additional parameters.
Subsequently, the video latent $z_{0}^{v}$ is noised according to the flow-matching defined in Eq.\,\ref{eq:1}, while the reference latent $z^{\text{src}}_{0}$ and garment latent $z^{\text{gar}}_{0}$ remain noise-free as conditional inputs.

\noindent
\textbf{Multi-Modal Attention.}
To enable multi-modal interaction within a single backbone, the clean reference latent $\mathbf{z}^{\text{src}}_{0}$, clean garment latent $\mathbf{z}^{\text{gar}}_{0}$, and noisy video latent $\mathbf{z}^{v}_{t}$ are concatenated along the token dimension. The resulting sequence $z_t^{\text{uni}}$ is then projected via learnable matrices $W_q$, $W_k$, and $W_v$, followed by multi-modal attention interaction. The attention output $\mathcal{O}$ can be formulated by:
\begin{equation}
    \mathcal{O} = \text{Softmax}(\frac{(\mathcal{W}_q \cdot z_t^{\text{uni}} )(\mathcal{W}_k \cdot z_t^{\text{uni}})^\top}{\sqrt{d_k}})(\mathcal{W}_v \cdot z_t^{\text{uni}}),
\end{equation}
where $d_k$ denotes the feature dimension.
These shared projection matrices enables global interaction between conditional and video latents without introducing additional parameters.
Finally, the model output retains only the video latent, discarding the reference latent and garment latent.

\subsection{Streaming Distillation with In-Context Learning}
\label{sec:3-2}
In this section, we distill the pretrained teacher into a few-step autoregressive student for streaming generation.
Prior works~\cite{yin2025slow,huang2025self,zhu2026causal} show that direct distillation is challenging and adopt a two-stage strategy comprising ODE initialization and distribution matching distillation~\cite{yin2024improved}.
To better adapt to our setting, we instead adopt \textit{teacher forcing}~\cite{gao2024ca2,hu2024acdit,zhang2025test} to initialize the student model, followed by \textit{gradient-reweighted distribution matching distillation} to improve extrapolation consistency.

\noindent
\textbf{In-Context Teacher Forcing Mask.}
The teacher forcing fine-tunes the pretrained multi-step bidirectional model into a multi-step autoregressive model using clean data.
However, unlike prior approaches~\cite{huang2025live,zhuang2025flashvsr,shin2025motionstream} that inject control signals via adapters, our model incorporates these signals through in-context token concatenation, making standard teacher forcing inapplicable.
To address this, we design an in-context teacher forcing mask for training, with the toy examples shown in Figure~\ref{fig:overall_pipeline}.
Specifically, in addition to the noisy sequence $\langle z^{\text{src}}_0, z^{\text{tar}}_0, z^{v}_t \rangle$, we symmetrically concatenate its clean counterpart $\langle z^{\text{src}}_0, z^{\text{tar}}_0, z^{v}_0 \rangle$ and feed the resulting sequence into the model.
For the conditioning signals $z^{\text{src}}_0$ and $z^{\text{tar}}_0$, we apply a dedicated masking strategy such that all generated frames can attend to them, while $z^{\text{src}}_0$ and $z^{\text{tar}}_0$ cannot access any future generated frames.
In this way, when predicting the next frame (chunk), model conditions on ground-truth historical frames and conditional signals.

\noindent
\textbf{Gradient-Reweighted Distribution Matching Distillation.}
Based on the autoregressive model fine-tuned with teacher forcing, we further apply distribution matching distillation (DMD) for few-step generation and combine it with Self-Forcing~\cite{huang2025self} to better align training with inference.
However, we observe that directly applying DMD often leads to distorted human motions during extrapolation.
We attribute this to the unequal difficulty of frames in self-rolling generation: errors accumulate over time, making later frames more prone to drift, whereas vanilla DMD weights all frames equally.
To resolve this, we propose an adaptive gradient reweighting strategy that increases the weights of low-quality frames while decreasing those of high-quality ones during distillation.
Specifically, we use an aesthetic reward model $\mathcal{R}$ to estimate frame quality during distillation and normalize the resulting scores into frame-wise gradient weights.
In this way, the Eq.\,\ref{eq:3} can be rewritten as:
\begin{equation}
\begin{gathered}
\nabla \mathcal{L}_{\text{Reweight-DMD}} 
= - \mathbb{E}_{t} \Biggl[
\int 
\mathcal{A}^{1:f} (G(\epsilon)) 
\cdot
\big(
s_{\text{real}}^{1:f} (\phi(G(\epsilon), t), t)
-
s_{\text{fake}}^{1:f} (\phi(G(\epsilon), t), t)
\big)
\cdot
\frac{d G_{\theta}(\epsilon)}{d\theta}
\cdot
d\epsilon
\Biggr], \\
\mathcal{A}^{i} (G(\epsilon)) = \frac{\exp(-\mathcal{R} (G^{i}(\epsilon)) / \tau)}{\sum_{j=1}^{f} \exp(-\mathcal{R} (G^{j}(\epsilon)) / \tau)}, \quad i=1, \dots, f,
\end{gathered}
\end{equation}
where $\tau$ denotes the temperature coefficient that controls the relative weight. Note that this approach is not restricted to aesthetic rewards and can naturally accommodate other reward models.

\subsection{Training-Free KV Cache Rescheduling}
\label{sec:3-3}
Given the distilled few-step autoregressive models, we manage KV cache to enable stable long-video extrapolation.
In detail, the reference KV entry $KV^{\text{src}}$ and garment KV entry $KV^{\text{gar}}$ are persistently stored in the KV cache as conditioning signals.
Following prior work~\cite{yang2025longlive,yesiltepe2025infinity}, we also retain the KV entries of the initial frame (chunk), $KV^{0}$, as an attention sink to improve stability during extrapolation.
All remaining KV entries follow a first-in and first-out policy when the cache exceeds its maximum size.
Formally, at the generation of $\text{k-th}$ frame, the KV cache is defined as:
\begin{equation}
    \text{KV Cache} \, := \, \langle KV^{\text{src}}, KV^{\text{gar}}, KV^{0}, KV^{\text{Max(1, k - M + 4)}}, \dots, KV^{\text{k}} \rangle,
\end{equation}
where $M$ is the maximum KV cache size.
To enable \textbf{interactive multi-garment switching while maintaining coherence}, we reschedule the KV cache via three mechanisms: \textit{Garment KV Refresh}, \textit{Historical KV Withdraw}, and \textit{Reference KV Disentangle}, as illustrated in Figure~\ref{fig:overall_pipeline} (right).

\begin{figure}[t]
    \centering
    \includegraphics[width=0.98\linewidth]{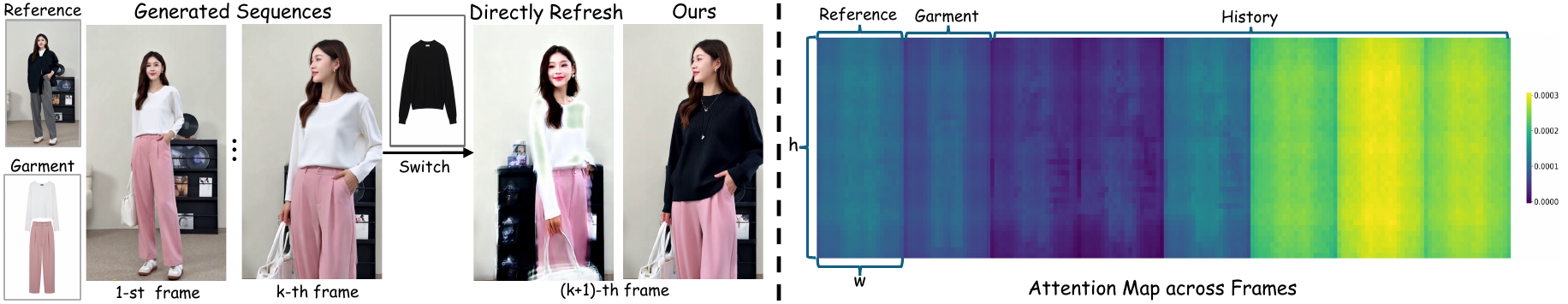}
    \caption{
    (Left) Generated sequences during garment switching. Directly refreshing the garment KV fails to change the subject’s clothing, while our KV cache rescheduling enables garment-switching and motion coherence.
    (Right) Average attention visualization of newly generated frames over historical and conditional KV. The model attends more to historical KV than to conditional KV.
    }
    \label{fig:analysis}
    \vspace{-1.0em}
\end{figure}

\noindent
\textbf{Garment KV Refresh.}
To switch the character with a new garment $I^{\text{gar}_2}$ during generation, we refresh the garment KV in the cache.
Specifically, $I^{\text{gar}_2}$ is encoded into $z^{\text{gar}_2}$ by VAE, and the corresponding  $KV^{\text{gar}_{2}}$ are obtained via a forward pass. 
We then replace the old $KV^{\text{gar}}$ in the cache with new new $KV^{\text{gar}_{2}}$, so that subsequent frames are generated conditioned on the updated garment.

\noindent
\textbf{Historical KV Withdraw.}
However, as shown in Figure~\ref{fig:analysis} (left), directly refreshing garment KV is insufficient to change the garment in subsequent generated frames.
To analyze this phenomenon, we visualize the average attention weights of newly generated latents over conditional and historical KV. In Figure~\ref{fig:analysis} (right), attention is more concentrated on historical KV rather than conditional KV.
This indicates that, under streaming eneration with in-context learning, the model relies more on historical context than on conditional signals.
Consequently, the old garment from historical frames tends to persist in newly generated frames, rendering the new garment signal ineffective.
Therefore, we withdraw the historical KV, encouraging the model to focus on the new garment KV.

\noindent
\textbf{Reference KV Disentangle.}
While withdrawing historical KV enables garment switching, it weakens temporal coherence across the switching frame.
Recall that we deliberately \textbf{I2V property} during pre-training, in which the first generated frame remains consistent with the reference frame except for garment information. This endows the model with an implicit capability to maintain temporal coherence during single-garment switching.
To enable multi-garment switching during generation, the key is to \textit{align the distribution of the new conditioning signal with that of the original conditioning signal}.
To this end, we replace old $KV^{\text{src}}$ with the $KV^{\text{k}}$ extracted from the last historical frame.
\textbf{Notably}, the new reference KV corresponds to four decoded frames, mismatching with the old reference KV that corresponds to single-frame.
We thus perform a VAE decode-encode process to disentangle the last decoded frame, followed by an additional forward to obtain new reference KV.

\subsection{High-Quality Data Curation Pipeline}
\label{sec:3-4}
To further support teacher model pre-training and streaming distillation post-training, we design a data curation pipeline to construct samples of the reference image $I^{\text{src}}$, garment image $I^{\text{gar}}$, video sequence $\mathcal{V}$ and corresponding prompt.
The pipeline consists of four stages: \textit{1. General Coarse-to-Fine Video Filtering}, \textit{2. Static-Dynamic Video Captioning}, \textit{3. Fine-Grained Garment Images Extraction}, and \textit{4. Adaptive Reference Images Construction}.
We provide implementation details in the \textbf{Appendix}.

%% file: sections/4-exp.tex
\section{Experiments}

\subsection{Experimental Details.}

\noindent
\textbf{Implementation Details.}
Our teacher model is initialized with WAN2.2-5B-TI2V~\cite{wan2025wan}.
During streaming distillation, we use an aesthetic scorer as the reward model, with the temperature coefficient $\tau$ set to $0.2$.
During inference, the KV cache size $M=23$. We adopt a chunk-wise generation strategy, where each chunk consists of $3$ latent frames.
All experiments are conducted on NVIDIA A100 GPUs.
Due to space limitations, we provide additional training details in the \textbf{Appendix}.

\noindent
\textbf{Evaluation Settings.}
The task most closely related to ours is multi-reference customized video generation. Accordingly, we select several representative baselines: VACE~\cite{vace}, Kaleido~\cite{zhang2025kaleido}, MAGREF~\cite{deng2025magref}, SkyReels-A2~\cite{fei2025skyreels} and Phantom~\cite{liu2025phantom}.
Moreover, we compare with a first-frame editing + Image-to-Video (I2V) pipeline, where Qwen-Image-Edit~\cite{wu2025qwen} performs editing, followed by WAN-5B-TI2V~\cite{wan2025wan} for I2V generation.
Note that all baselines generate videos at their respective native resolutions and durations.
To evaluate different methods on the human-garment video customization task, we construct a benchmark termed HGC-Bench.
HGC-Bench contains $240$ samples, each consisting of a reference character image, a garment image, and a corresponding prompt, covering a wide range of characters, scenarios, and garments. We provide additional details in the \textbf{Appendix}.

\begin{table}[!t]
\centering
\caption{
Quantitative comparison of different methods for \textit{short ($81$ frames) video customized generation}. The best results are highlighted in \textbf{bold} and the second best are \uline{underlined}. Note that the frames per second (FPS) of all methods are evaluated on an H200 GPU.
}
\label{tab:main_results}
\small
\setlength{\tabcolsep}{2.5pt}
\begin{tabular}{l c ccccccccc}
\toprule
Methods & Params $\downarrow$ & Cur. $\uparrow$ & GME $\uparrow$ & Amp. $\uparrow$ & Smoo. $\uparrow$ & VQ $\uparrow$ & HGC $\uparrow$ & LGC $\uparrow$ & NTP $\uparrow$ & FPS $\uparrow$ \\
\midrule
Edit~\cite{wu2025qwen}+I2V~\cite{wan2025wan} & 20B+5B & 0.4094 & 0.6741 & \textbf{0.8636} & 0.9898 & \uline{0.7482} & \uline{4.5417} & \uline{3.9167} & 4.4583 & 0.76 \\
VACE~\cite{vace}                             & 14B & 0.2746 & 0.6962 & 0.4054 & 0.9764 & 0.7409 & 4.3708 & 3.5458 & 4.6417 & 0.23 \\
Kaleido~\cite{zhang2025kaleido}              & 14B & 0.3676 & 0.6882 & 0.2675 & \uline{0.9935} & 0.7478 & 4.1708 & 3.5500 & \uline{4.7167} & 0.13 \\
MAGREF~\cite{deng2025magref}                 & 14B & 0.0459 & \textbf{0.7138} & 0.2571 & 0.9436 & 0.7301 & 3.6000 & 2.2000 & 2.6875 & 0.27 \\
SkyReels-A2~\cite{fei2025skyreels}           & 14B & 0.3689 & 0.6550 & 0.5205 & 0.9424 & 0.7241 & 3.3625 &  2.6958 & 4.6458 & 0.54 \\
Phantom~\cite{liu2025phantom}                & 1.3B & \textbf{0.5507} & 0.6855 & 0.1144 & 0.9668 & 0.7338 & 4.3292 & 3.6417 & 4.6875 & \uline{0.77} \\
Phantom ~\cite{liu2025phantom}               & 14B & \uline{0.4911} & \uline{0.6972} & 0.2086 & 0.9932 & 0.7446 & 4.5375 & 3.8333 & 4.6417 & 0.15 \\
\rowcolor{gray!25}
\method     & 5B & \uline{0.4911} & 0.6839 & \uline{0.7771} & \textbf{0.9969} & \textbf{0.7483} & \textbf{4.6833} & \textbf{3.9250} & \textbf{4.7625} & \textbf{23.8} \\
\bottomrule
\end{tabular}
\end{table}

\begin{figure}[t]
    \centering
    \includegraphics[width=0.98\linewidth]{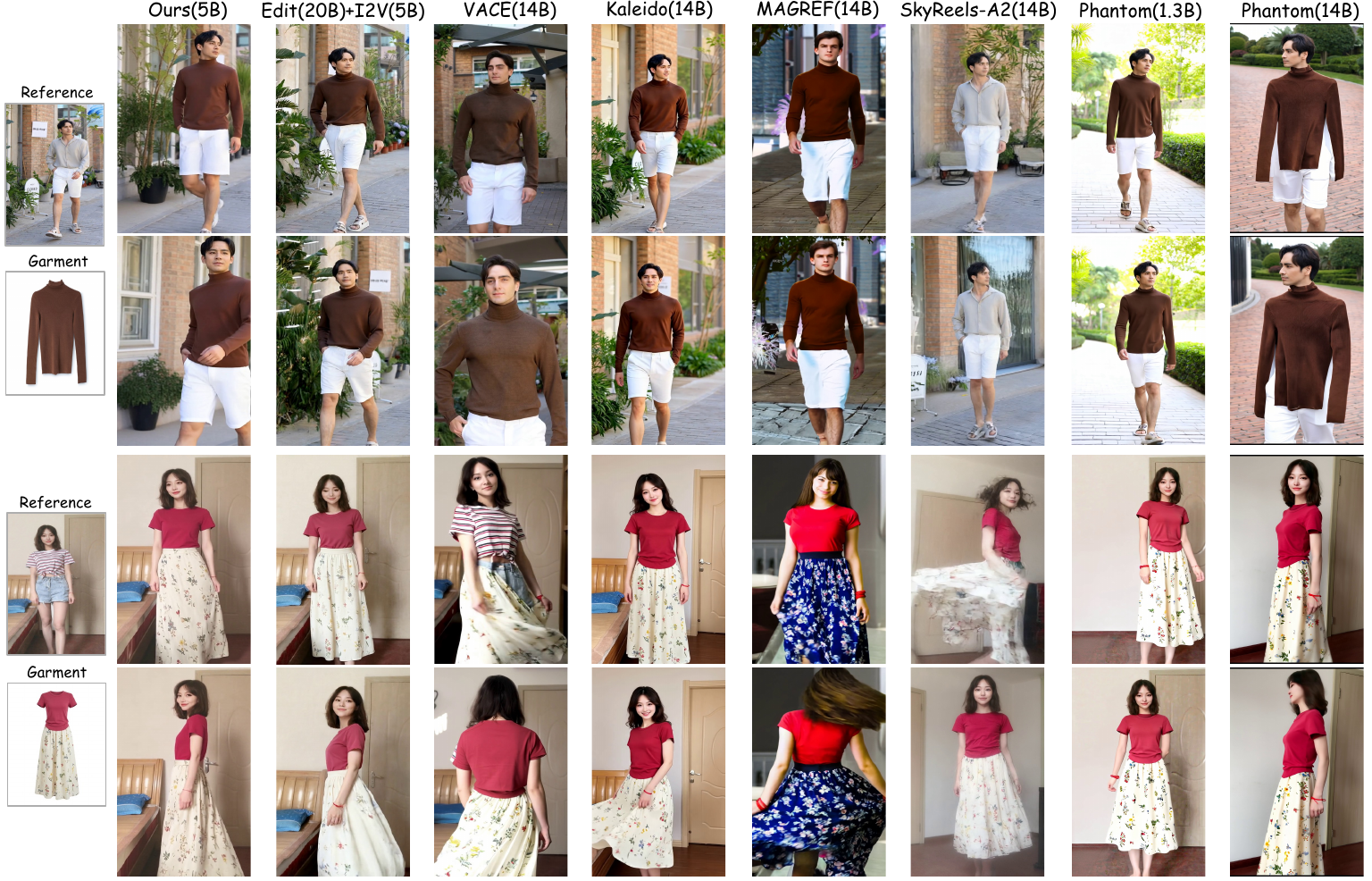}
    \caption{
    Qualitative comparison of our \method with other baselines.
    Due to space limitations, we omit the input prompts here; please refer to the Appendix for details.
    }
    \label{fig:qualitative_comparison}
    \vspace{-1.0em}
\end{figure}

\subsection{Main Results}

\noindent
\textbf{Quantitative Comparisons.}
Inspired by prior works~\cite{deng2025magref,liu2025phantom,xue2025stand}, we adopt several evaluation metrics, including ID consistency (Cur Score), text alignment (GME Score), motion magnitude (Amplitude), and temporal smoothness (Smoothness) following OpenS2V-Nexus~\cite{yuan2025opens2v}, as well as overall visual quality (VQ Score) following VBench~\cite{huang2024vbench}.
To assess garment consistency, we use Gemini-3.0 to evaluate the generated results from three aspects: high-level garment consistency (HGC), low-level garment consistency (LGC), and non-target garment preservation (NTP).
In addition, we report the frames per second (FPS) of each method to measure efficiency. See \textbf{Appendix} for details.
In Table~\ref{tab:main_results}, \method outperforms all baselines in temporal consistency, video quality, and three garment consistency metrics.
For ID consistency and motion magnitude, our method ranks second, following the Phantom(1.3B)~\cite{liu2025phantom} and Edit~\cite{wu2025qwen}+I2V~\cite{wan2025wan}, respectively.
\textbf{Notably}, \method significantly outperforms all baselines in efficiency, enabling real-time generation at $23.8$ FPS.

\begin{figure}[t]
    \centering
    \includegraphics[width=0.98\linewidth]{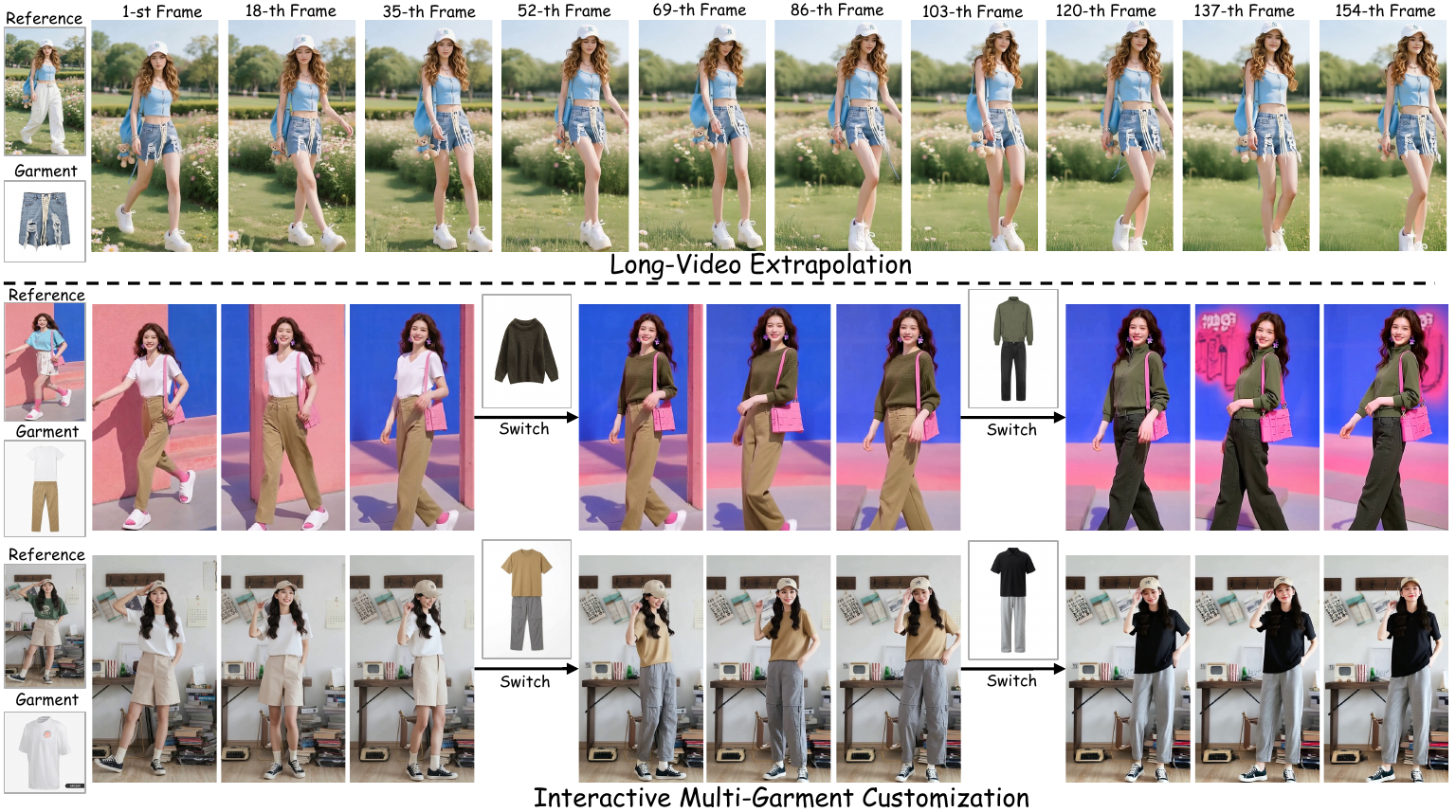}
    \caption{
    Additional applications of \method. It supports both \textit{long-video extrapolation} and \textit{interactive multi-garment customization}.
    We omit prompts for brevity; see Appendix for details.
    }
    \label{fig:app}
    \vspace{-1.0em}
\end{figure}

\noindent
\textbf{Qualitative Comparisons.}
We further provide qualitative comparisons to assess ID consistency, garment consistency, and overall visual fidelity across different methods. As shown in Figure~\ref{fig:qualitative_comparison}, existing approaches often struggle to simultaneously maintain subject identity, garment details, and natural motions.
In cases involving large pose variations or with complex garments, these methods tend to exhibit noticeable degradation in appearance and garment preservation.
Moreover, several baselines exhibit garment mismatch or unintended modifications to non-target garments, which degrade overall realism and temporal consistency across frames. See \textbf{Appendix} for more results.

\noindent
\textbf{Long-Video Extrapolation.}
Existing multi-reference customization methods rely on bidirectional architectures that synthesize all frames jointly, making them unsuitable for long-video customized generation.
In contrast, the autoregressive generation paradigm of \method naturally supports long-video extrapolation.
As shown in Figure~\ref{fig:app}, \method can maintain character consistency and garment consistency across long temporal ranges. See \textbf{Appendix} for more results.

\noindent
\textbf{Interactive Customization.}
Benefiting from proposed KV Cache Rescheduling, \method further enables interactive multi-garment customized generation, which is beyond the capability of existing methods.
As shown in Figure~\ref{fig:app}, \method supports interactive garment-switching during generation while preserving coherent human motion. See \textbf{Appendix} for more results.

\begin{figure}[t]
    \centering
    \includegraphics[width=0.98\linewidth]{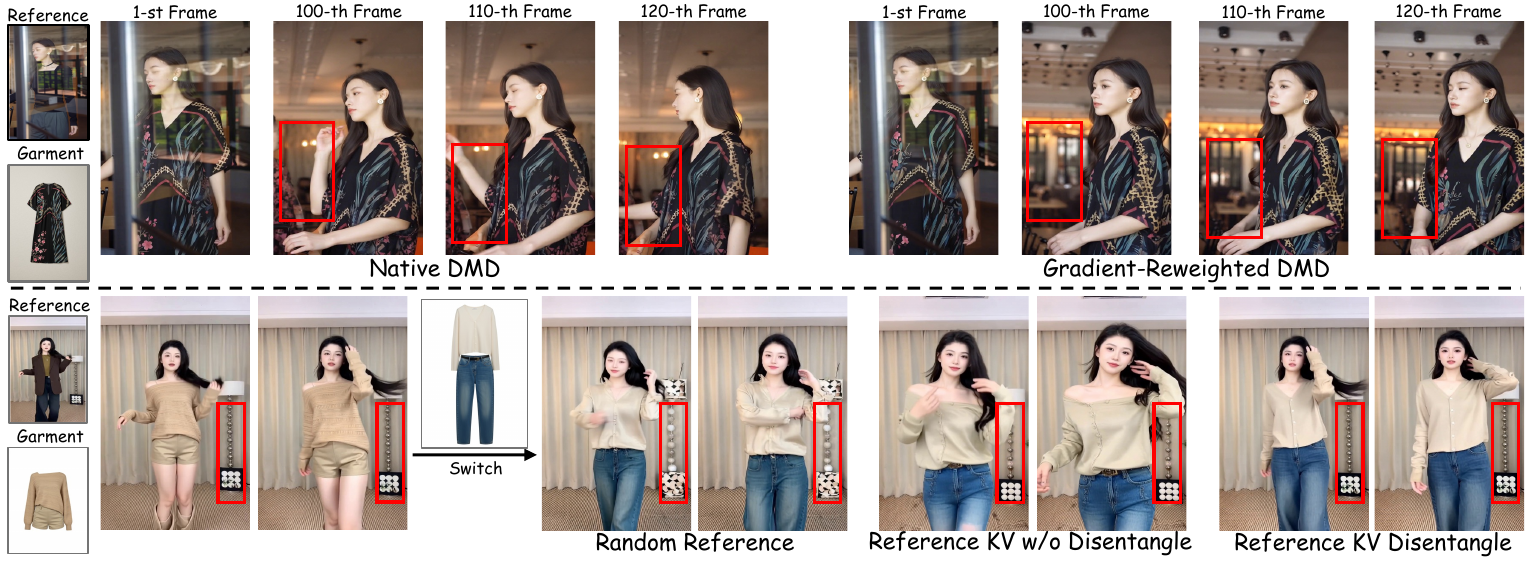}
    \caption{
    Qualitative ablation of \textit{Gradient-Reweighted Distribution Matching Distillation (DMD)} and \textit{Reference KV Disentangle}. 
    Gradient-Reweighted DMD alleviates motion collapse during extrapolation, while Reference KV Disentangle further enhances consistency during garment switching.
    }
    \label{fig:ablation2}
    \vspace{-1.0em}
\end{figure}

\begin{table}[!t]
\centering
\caption{
Quantitative ablation of teacher training strategies for \textit{short ($81$ frames) video customized generation}. 
The best results are highlighted in \textbf{bold} and the second best are \uline{underlined}.
}
\setlength{\tabcolsep}{3.5pt}
\begin{tabular}{l c ccccccccc}
\toprule
Variants & Cur. $\uparrow$ & GME $\uparrow$ & Amp. $\uparrow$ & Smoo. $\uparrow$ & VQ $\uparrow$ & HGC $\uparrow$ & LGC $\uparrow$ & NTP $\uparrow$ \\
\midrule
Chan.-Concat + Full FT     & 0.1811 & 0.6874 & 0.3748 & 0.9266 & 0.7404 & 4.4917 & 3.1667 & 4.4667 \\
\rowcolor{gray!25}
Ours + Full FT             & \textbf{0.4602} & \textbf{0.6972} & 0.5625 & \textbf{0.9936} & \textbf{0.7473} & \textbf{4.8583} & \textbf{4.1583} & \textbf{4.7792} \\
Ours + Attn FT             & \uline{0.4348} & \uline{0.6900} & \uline{0.6350} & \uline{0.9881} & \uline{0.7471} & \uline{4.8500} & \uline{4.0625} & \uline{4.7750} \\
Ours + LoRA~\cite{hu2022lora} FT               & 0.4046 & 0.6928 & \textbf{0.6448} & 0.9777 & 0.7437 & 4.7292 & 3.9458 & 4.7042 \\
\bottomrule
\label{tab:ablation1}
\vspace{-1.0em}
\end{tabular}
\end{table}

\begin{table}[!t]
\centering
\caption{
Quantitative ablation of Gradient-Reweighted Distribution Matching Distillation (GR-DMD) for \textit{long ($165$ frames) video customized generation}. The best results are highlighted in \textbf{bold}.
}
\setlength{\tabcolsep}{2.5pt}
\begin{tabular}{l c ccccccccc}
\toprule
Variants & Cur. $\uparrow$ & GME $\uparrow$ & Amp. $\uparrow$ & Smoo. $\uparrow$ & VQ $\uparrow$ & HGC $\uparrow$ & LGC $\uparrow$ & NTP $\uparrow$ \\
\midrule
Naive DMD           & 0.4232 & 0.6700 & 0.8026 & 0.9932 & 0.7419 & 4.6958 & 3.8958 & 4.7125  \\
\rowcolor{gray!25}
GR-DMD ($\tau=0.2$) & \textbf{0.4265} & 0.6732 & \textbf{0.8395} & \textbf{0.9975} & \textbf{0.7480} & 4.7000 & 3.9042 & \textbf{4.7333}  \\
GR-DMD ($\tau=0.3$) & 0.4111 & \textbf{0.6786} & 0.5106 & 0.9933 & 0.7465 & \textbf{4.7583} & \textbf{3.9375} & 4.6958  \\
GR-DMD ($\tau=0.4$) & 0.4047 & 0.6696 & 0.7869 & 0.9872 & 0.7424 & 4.7125 & 3.9022 & 4.7208  \\
GR-DMD ($\tau=0.5$) & 0.4252 & 0.6774 & 0.7907 & 0.9953 & 0.7421 & 4.7083 & 3.8833 & 4.7058  \\
\bottomrule
\label{tab:ablation2}
\vspace{-1.0em}
\end{tabular}
\end{table}

\subsection{Ablation Studies}
In this section, we conduct three groups of ablation studies: \textit{Teacher Model}, \textit{Streaming Distillation}, and \textit{KV Cache Rescheduling}.
Additional ablation results are provided in the \textbf{Appendix}.

\noindent
\textbf{Ablation with Teacher Model.}
To validate the effectiveness of \textit{In-Context Learning}, we compare it with channel-wise concatenation.
In Table~\ref{tab:ablation1}, our designed in-context learning outperforms simple channel-wise concatenation across several metrics.
Moreover, we compare different fine-tuning (FT) strategies, including Full FT, Attn FT, and LoRA~\cite{hu2022lora} FT, with the results shown in Table~\ref{tab:ablation1}.
Full FT performs best overall, so we adopt this version of the teacher model for streaming distillation.

\noindent
\textbf{Ablation with Streaming Distillation.}
We first analyze the effectiveness of \textit{Gradient-Reweighted Distribution Matching Distillation (GR-DMD)} in long-video (165 frames) extrapolation through qualitative and quantitative evaluations, as shown in Table~\ref{tab:ablation2} and Figure~\ref{fig:ablation2}.
Intuitively, naive DMD tends to produce distorted or duplicated human limbs during extrapolation.
In contrast, our Gradient-Reweighted DMD generates coherent and anatomically consistent human structures during extrapolation.
Moreover, we further investigate the effect of the temperature coefficient $\tau$ on long-video extrapolation.
In Table~\ref{tab:ablation2}, the hyper-parameter $\tau=0.2$ yields the best overall performance.

\noindent
\textbf{Ablation with KV Cache Rescheduling.}
We now analyze the choice of reference KV and the effectiveness of disentanglement, as visualized in Figure~\ref{fig:ablation2}.
Clearly, randomly selecting reference KV leads to inconsistencies with previous frames.
This phenomenon stems from the image-to-video prior, where the generated initial frame aligns with the reference image; thus, mismatched reference KV breaks temporal coherence.
Moreover, without disentangling the last historical KV, distribution mismatch arises: the reference frame is independently VAE-encoded during training, while the non-disentangled historical KV corresponds to multiple decoded frames (\emph{e.g.}, four).

%% file: sections/5-conclusion.tex
\section{Conclusion}
In conclusion, we present \textbf{\method}, a real-time and interactive framework for human-garment customization in autoregressive video generation, where users can interactively switch garment during generation.
\method consists of three key techniques:
(i) We develop a \textit{Teacher Model with In-Context Learning}  to encourage the model to implicitly preserve coherence during single-garment switching.
(ii) We introduce \textit{Streaming Distillation with In-Context Learning} to enable efficient inference and consistent long-video extrapolation.
(iii) We propose \textit{Training-Free KV Cache Rescheduling} to support interactive multi-garment video customization while preserving coherent human motion.
Extensive experiments show that our \method demonstrates superiority over existing approaches while achieving real-time 720p video generation at 23.8 fps on a single GPU. 
Additional experiments on interactive customization and long-video extrapolation showcase its practical value in human-centric applications such as e-commerce and content creation.

%% file: sections/X-suppl.tex
\appendix

\begin{figure}[t]
    \centering
    \includegraphics[width=0.98\linewidth]{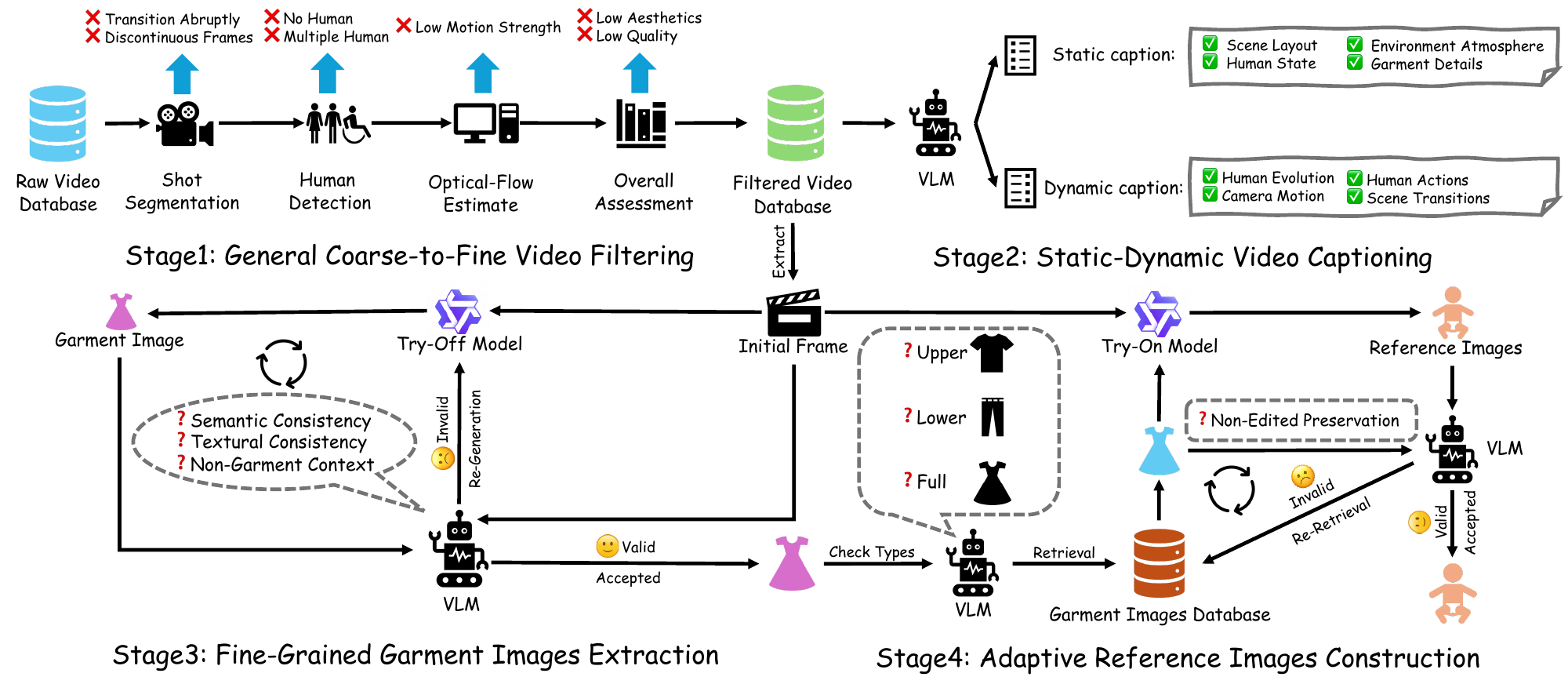}
    \caption{
    The high-quality data curation pipeline of \method. It consists of four stages: \textit{(1) General Coarse-to-Fine Video Filtering}, \textit{(2) Static-Dynamic Video Captioning}, \textit{(3) Fine-Grained Garment Image Extraction}, and \textit{(4) Adaptive Reference Image Construction}.
    }
    \label{fig:data_pipeline}
\end{figure}

\section{Data Curation Pipeline Details}
Recall that we briefly introduce our high-quality data curation pipeline in the main paper, which comprises four stages: \textit{1. General Coarse-to-Fine Video Filtering}, \textit{2. Static-Dynamic Video Captioning}, \textit{3. Fine-Grained Garment Image Extraction}, and \textit{4. Adaptive Reference Image Construction}.
The overall curation pipeline is illustrated in Figure~\ref{fig:data_pipeline}, and we detail each stage as follows:

\noindent
\textbf{1. General Coarse-to-Fine Video Filtering.}
We collected a large set of raw videos from the Internet and filtered them in a coarse-to-fine manner using \textit{Shot Segmentation}, \textit{Human Detection}, \textit{Optical-Flow Estimation}, and \textit{Overall Assessment} to retain only qualified videos:
\begin{itemize}
    \item \textbf{Shot Segmentation.} The raw videos are first processed with PySceneDetect to identify scene transitions and split into separate scene clips. These clips are then further divided into 3-5 second subclips, while discontinuous or overly short subclips are removed.
    \item \textbf{Human Detection.} We apply YOLOv8-Seg to each subclip to detect human presence and retain only single-person clips. Clips without humans or with multiple prominent humans are removed. Note that a clip is still considered single-person if one person occupies most of the frame and any other visible people appear only as small, blurred background figures.
    \item \textbf{Optical-Flow Estimation} For each subclip containing one human, we estimate optical flow using UniMatch~\cite{li2025unimatch} to measure motion magnitude. We then retain clips with moderate to large motion and discard clips with little or slow motion based on a predefined threshold.
    \item \textbf{Overall Assessment.} Finally, we evaluate each subclip using Q-Align~\cite{wu2023q} for aesthetics and FAST-VQA-M~\cite{wu2022fast} for overall visual quality. We retain clips with high aesthetic and quality scores according to predefined thresholds, and remove those with low scores.
\end{itemize}

\noindent
\textbf{2. Static-Dynamic Video Captioning.}
For the filtered videos, we use the vision-language model (VLM) Gemini-3.1 to generate captions. Specifically, we adopt a static-dynamic decoupling strategy:
\begin{itemize}
    \item \textbf{Static Caption.} We prompt the VLM to focus on the static content in each video, including the scene layout, environmental atmosphere, human attributes (\emph{e.g.}, appearance), and garment details. These elements are intrinsic to the video and remain unchanged over time.
    \item \textbf{Dynamic Caption.} We then prompt the VLM to capture the dynamic content of each video, including human evolution (\emph{e.g}. facial expressions), human action, camera motion, and scene transitions. These elements are inherently temporal and typically change over time.
\end{itemize}
The system prompt for Gemini-3.1 is presented in Sec.\,\ref{sec:system_prompt}.

\noindent
\textbf{3. Fine-Grained Garment Images Extraction.}
For each filtered video, we extract the initial frame and apply the image try-off model Qwen-Image-Edit~\cite{wu2025qwen} to extract corresponding garment images.
Since try-off is not always reliable in practice, we further introduce a VLM to verify the extracted garments.
In detail, for each extracted garment, the VLM performs a three-stage validity check:
\begin{itemize}
\item \textbf{Semantic Consistency.} The VLM will check whether the extracted garment matches the clothing in the initial frame at a high level, such as garment category and color.
\item \textbf{Textural Consistency.} The VLM will check whether the extracted garment matches the clothing in the initial frame at a low level, such as texture and logos.
\item \textbf{Non-Garment Context.} The VLM will check whether the extracted garment contains information beyond the garment itself, such as irrelevant scene content or other artifacts.
\end{itemize}
We reapply the image try-off model until the extracted result passes all VLM-based validity checks. If extraction fails repeatedly, we discard the corresponding sample.

\noindent
\textbf{4. Adaptive Reference Images Construction.}
In the final stage, we construct the reference image. To improve training robustness, the garment worn by the person in the reference image should differ from the extracted garment.
We note that the garment information extracted in the previous stage may be incomplete, for example, including only the upper-body or lower-body clothing.
To fully utilize the available garment information, we employ the VLM Gemini-3.1 to guide the accurate construction of the reference image. In detail, the overall process is formulated as follows:
\begin{itemize}
\item \textbf{Garment Type Classification.} For the garment extracted from each video, the VLM first determines whether it corresponds to upper-body, lower-body, or full-body clothing.
\item \textbf{Garment Type Retrieval.} Based on the predicted garment category, the VLM will retrieve a visually compatible garment of the same type from the garment database.
\item \textbf{Accurate Image Try-On.} Given the retrieved garment and the extracted first frame, we apply an image try-on model to construct the reference image. This enables fine-grained customization, where the specified garment is changed while other regions remain unchanged.
\item \textbf{Validity Check.} We use a VLM to verify each reference image by checking whether the non-edited regions remain unchanged. If not, we reconstruct the reference image using the image try-on mode. We discard the corresponding sample if reconstruction fails repeatedly.
\end{itemize}

In total, we curate about 82K triplets, each consisting of a reference image, a garment image, and the corresponding video. After manual verification, about 62K triplets are retained in the training dataset.

\section{Training Details}
\noindent
\textbf{Pre-training Configuration.}
During teacher model pre-training, we keep the VAE in \texttt{float32} precision and fully fine-tune the transformer in \texttt{bfloat16}. 
To further improve GPU utilization, we adopt a Fully Sharded Data Parallel (FSDP) training strategy with a global batch size of $64$.
We optimize the model using AdamW with $\beta_1=0.9$, $\beta_2=0.999$, and a weight decay of $0.01$. 
We further employ a learning rate schedule with a warm-up of $200$ steps, followed by a two-stage decay: the learning rate is set to $1\times10^{-5}$ until step $1100$ and then decayed to $5\times10^{-6}$ until step $2300$.

\noindent
\textbf{Post-training Configuration.}
During streaming distillation post-training, we maintain both the VAE and transformer in bfloat16 and also adopt FSDP training strategy with a global batch size of $64$.
\textit{For teacher forcing}, the generator is initialized from the pre-trained teacher model and then fully fine-tuned for $4000$ steps using AdamW with a learning rate of $1\times10^{-6}$, $\beta_1=0.0$, $\beta_2=0.999$, and a weight decay of $0.01$.
\textit{For gradient-reweighted distribution distillation matching}, the generator is initialized from the model fine-tuned with teacher forcing, while both the real score and fake score networks are initialized from the pre-trained teacher model. The few-step generator uses a timestep schedule of $[1000, 750, 500, 250]$. 
We fully fine-tune the generator and the fake score network with a 1:5 update ratio, while keeping the real score network frozen.
We optimize both generator and fake score network with AdamW for $400$ steps, using learning rates of $2\times10^{-6}$ for the generator and $4\times10^{-7}$ for the fake score network, with $\beta_1=0.0$, $\beta_2=0.999$, and a weight decay of $1\times10^{-2}$.

\noindent
\textbf{Dataset Configuration.}
For both pre-training and post-training, we use a carefully curated paired dataset of 62K samples, each consisting of a reference image, a garment image, and a video sequence. We sample sequences of $81$ frames to align with existing customization methods.
The video and reference image are resized to $1280 \times 704$ while preserving aspect ratio, whereas the garment image is center-padded to $1280 \times 704$ with aspect ratio preserved, following the standard resolution of WAN2.2-5B-TI2V~\cite{wan2025wan}.
\textit{During pre-training}, since the reference image already contains rich static information, we use only the dynamic content with a probability of 70\%, and use the full caption (static-dynamic contents), in the remaining 30\% of cases.
This encourages the model to infer static attributes directly from the reference image, reducing its reliance on textual descriptions.
\textit{During post-training}, we observe that using full captions, which include both static and dynamic content, leads to improved performance. We provide a more comprehensive analysis in Sec.\,~\ref{sec:ablation4}.
\textit{During interactive inference}, we intentionally avoid including garment-related descriptions in the input prompt, since the character's outfit is determined by the input garment image and may vary over time, which could otherwise conflict with fixed textual descriptions.

\begin{figure}[t]
    \centering
    \includegraphics[width=0.98\linewidth]{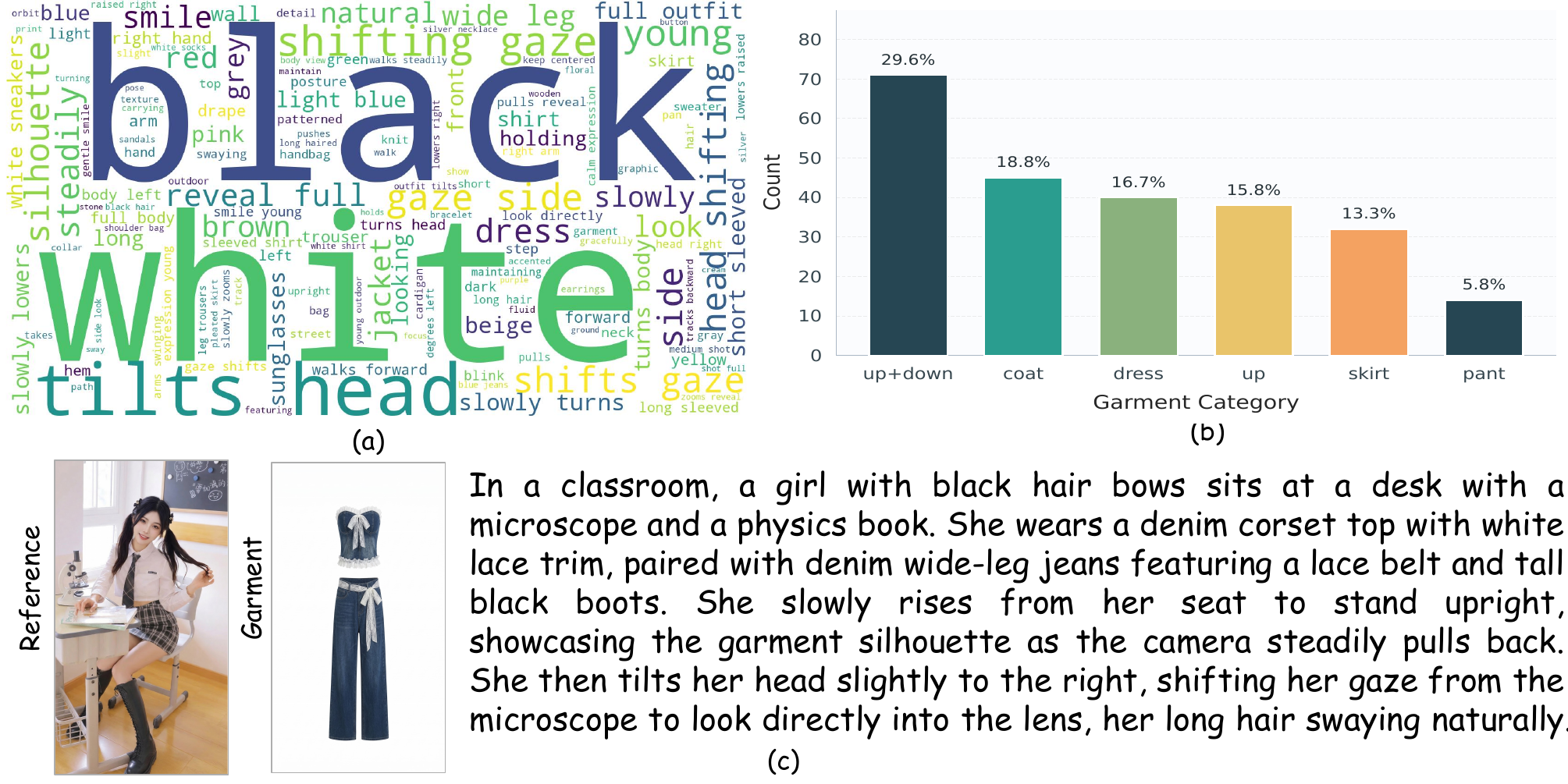}
    \caption{
    Data analysis and representative samples of HGC-Bench.
    (a) A word cloud generated from the input prompts, illustrating the diversity of scenarios and semantic content.
    (b) The distribution of garment categories, showing the proportions of different garment types.
    (c) Representative samples from HGC-Bench, each comprising a reference image, a garment image, and an input prompt.
    }
    \label{fig:hgc_bench}
    \vspace{-1.0em}
\end{figure}

\section{HGC-Bench Details}
We propose HGC-Bench, a dedicated benchmark for comprehensive evaluation. Specifically, we curate high aesthetic reference images from the Internet, anonymize identifiable facial information via face swapping, and pair them with corresponding garment images from our collected garment database.
Given the reference image and the garment image, we employ Gemini-3.0 to generate the corresponding prompt, which consists of a concise static description (\emph{e.g.}, human accessories and scene information), and a detailed dynamic description (\emph{e.g.}, human motions, camera movements).
In total, we curate $240$ samples, where each sample consists of a reference image, a garment image, and the corresponding prompt.
Figure~\ref{fig:hgc_bench} presents the data analysis and representative samples of HGC-Bench.
The system prompt for Gemini-3.0 is presented in Sec.\,\ref{sec:system_prompt}:

\section{Additional Ablation Studies on Distillation Prompts}
\label{sec:ablation4}
Recall that we adopted the hybrid caption strategy (70\% dynamic content and 30\% static-dynamic contents) during the teacher model training, to facilitate the extraction of static information from reference images.
During the streaming distillation (teacher forcing and gradient reweighted DMD) process, we find that using different types of captions can lead to different distilled results.
We quantify this effect, and the comparison results are reported in Table~\ref{tab:ablation3}.
Experimental results demonstrate that employing long caption (static-dynamic contents) yields superior performance.

\begin{table}[!t]
\centering
\caption{
Additional quantitative ablation on different distillation captions with $\tau=0.2$.
}
\setlength{\tabcolsep}{2.5pt}
\begin{tabular}{l c ccccccccc}
\toprule
Variants & Cur. $\uparrow$ & GME $\uparrow$ & Amp. $\uparrow$ & Smoo. $\uparrow$ & VQ $\uparrow$ & HGC $\uparrow$ & LGC $\uparrow$ & NTP $\uparrow$ \\
\midrule
Naive DMD (Mixed Caption)          & 0.4237 & 0.6564 & 0.7164 & 0.9797 & 0.7349 & 4.6234 & 3.8703 & 4.6444  \\
\rowcolor{gray!25}
Naive DMD (Long Caption)           & 0.4232 & 0.6700 & 0.8026 & 0.9932 & 0.7419 & 4.6958 & 3.8958 & 4.7125  \\
GR-DMD (Mixed Caption) & 0.4102 & 0.6692 & 1.1699 & 0.9955 & 0.7473 & 4.6583 & 3.9000 & 4.7369  \\
\rowcolor{gray!25}
GR-DMD (Long Caption) & 0.4265 & 0.6732 & 0.8395 & 0.9975 & 0.7480 & 4.7000 & 3.9042 & 4.7333  \\
\bottomrule
\label{tab:ablation3}
\end{tabular}
\vspace{-1.0em}
\end{table}

\section{Additional User Study}
\begin{figure}[t]
    \centering
    \includegraphics[width=0.98\linewidth]{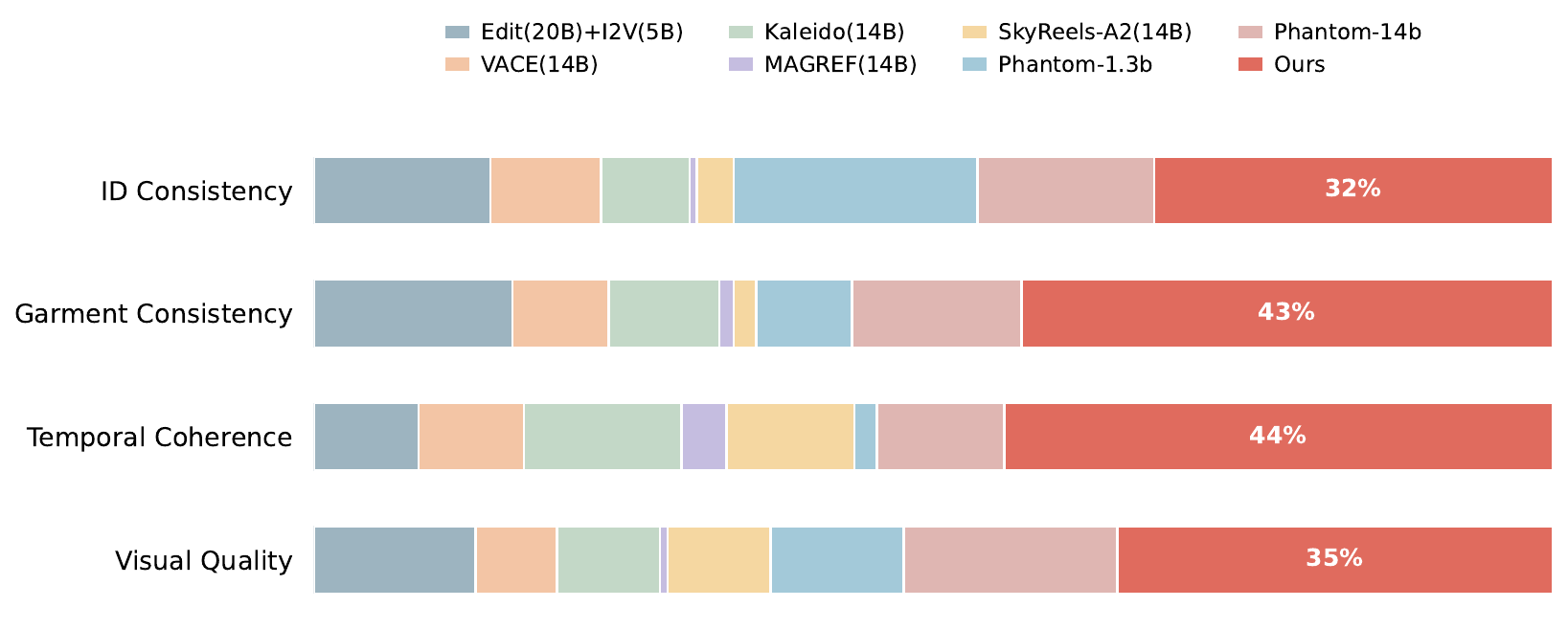}
    \caption{
    Quantitative results of the human evaluation. We compare our \method with other baselines across four key dimensions: \textit{ID Consistency}, \textit{Garment Consistency}, \textit{Temporal Coherence}, and \textit{Visual Quality}. Our \method achieves superior human preference rates.
    }
    \label{fig:user_study}
    \vspace{-1.0em}
\end{figure}

To evaluate user preference over videos generated by our method \method and other baselines, we conduct a user study.
In detail, for each comparison group, participants are shown videos generated by different methods and are asked to select the one with the best \textit{ID Consistency}, the best \textit{Garment Consistency}, the best \textit{Temporal Coherence}, and the best \textit{Visual Quality}.
In total, we collected $672$ valid responses, and the results are shown in Figure~\ref{fig:user_study}.
Our method achieves superior performance in id consistency, garment consistency, temporal coherence, and visual quality.

\section{Evaluation Details}
In this section, we provide a detailed clarification of the quantitative metrics used in the main paper.

\noindent
\textbf{ID Consistency (Cur Score)}
The Cur Score measures the consistency between the reference image and generated video.
Specifically, we extract facial embeddings from the reference image and each video frame using ArcFace~\cite{deng2019arcface} and compute the cosine similarity between the resulting embeddings.

\noindent
\textbf{Text Alignment (GME Score)}
The Gme Score is used to assess the semantic alignment between the generated video and the input prompt.
In detail, we utilize a vision-language model fine-tuned from Qwen2-VL~\cite{wang2024qwen2} to provide stronger capability in handling long and complex text descriptions.

\noindent
\textbf{Motion Magnitude (Amplitude)}
The Amplitude score measures motion amplitude in the generated video.
Specifically, we compute forward and backward optical flow between adjacent frames, calculate the flow magnitude, and average it over all pixels and frames to obtain the final score.

\noindent
\textbf{Temporal Smoothness (Smoothness)}
The Smoothness score evaluates the overall fluidity of motion in the generated video.
In particular, we utilize Q-Align~\cite{wu2023q} to measure the temporal coherence and the smoothness of motion transitions between consecutive frames.

\noindent
\textbf{Visual Quality (VQ Score)}
The VQ Score evaluates the overall visual quality of a video.
Specifically, we apply the no-reference image quality assessment model MUSIQ~\cite{ke2021musiq} to predict a quality score for each frame, and then average the frame-level scores to obtain the final video-level score.

\noindent
\textbf{Inference Efficiency (FPS)}
The frames per second (FPS) measures the inference efficiency of a model.
Specifically, we compute it as the total number of frames generated by the backbone network divided by the corresponding inference time.

\noindent
\textbf{Garment Consistency}
Besides the metrics above, we further evaluate the consistency between the garment worn by the character in the generated video and the given garment image.
As no established metric is available for this purpose, we employ the vision-language model Gemini-3.0 to assess this consistency from three dimensions: \textit{high-level garment consistency}, \textit{low-level garment consistency}, and \textit{non-target garment preservation}.
System prompt for Gemini-3.0 is provided in Sec.\,\ref{sec:system_prompt}.

\section{Limitations and Future Work}
While \method shows strong efficiency and interactivity in human-centric applications, several limitations remain: 
(i) Despite the curated data pipeline, the current training data still has limited garment categories and variations, which may restrict its generalization to complex scenarios.
(ii) The model remains challenged by complex human motions and camera movements, largely due to the imperfect performance of current open-source video generation backbones like Wan~\cite{wan2025wan}.

Therefore, future work could focus on developing a more efficient data curation pipeline, scaling up training datasets, and exploring stronger video generation backbones to address these limitations.

\section{Potential Negative Societal Impact}
Our \method is intended for human-garment customized video generation in human-centric content creation scenarios.
Nevertheless, we acknowledge that current models for human-garment video customization can introduce nontrivial societal risks when deployed irresponsibly or used with malicious intent.
We summarize our discussion in the following three points:
\begin{itemize}
    \item \textbf{Sexually Explicit or Violent Content.} Without proper safeguards, generated content may include sexually explicit, violent, or otherwise inappropriate material, potentially causing psychological or emotional harm to diverse audiences.
    \item \textbf{Stereotypes and Bias.} Unintended biases in character and garment information in the training data may be reflected or amplified in generated content, potentially reinforcing harmful cultural stereotypes or discriminatory visual representations.
    \item \textbf{Misleading Content.} Human-garment video customization models may be misused to create realistic yet false video advertisements, increasing the risk that misleading information spreads quickly and widely at scale.
\end{itemize}
We include these considerations to make clear that the method should be deployed responsibly and always accompanied by appropriate protections against misuse.

\section{Additional Qualitative Comparison}
\begin{figure*}[!t]
\begin{center}
   \includegraphics[width=0.98\textwidth] {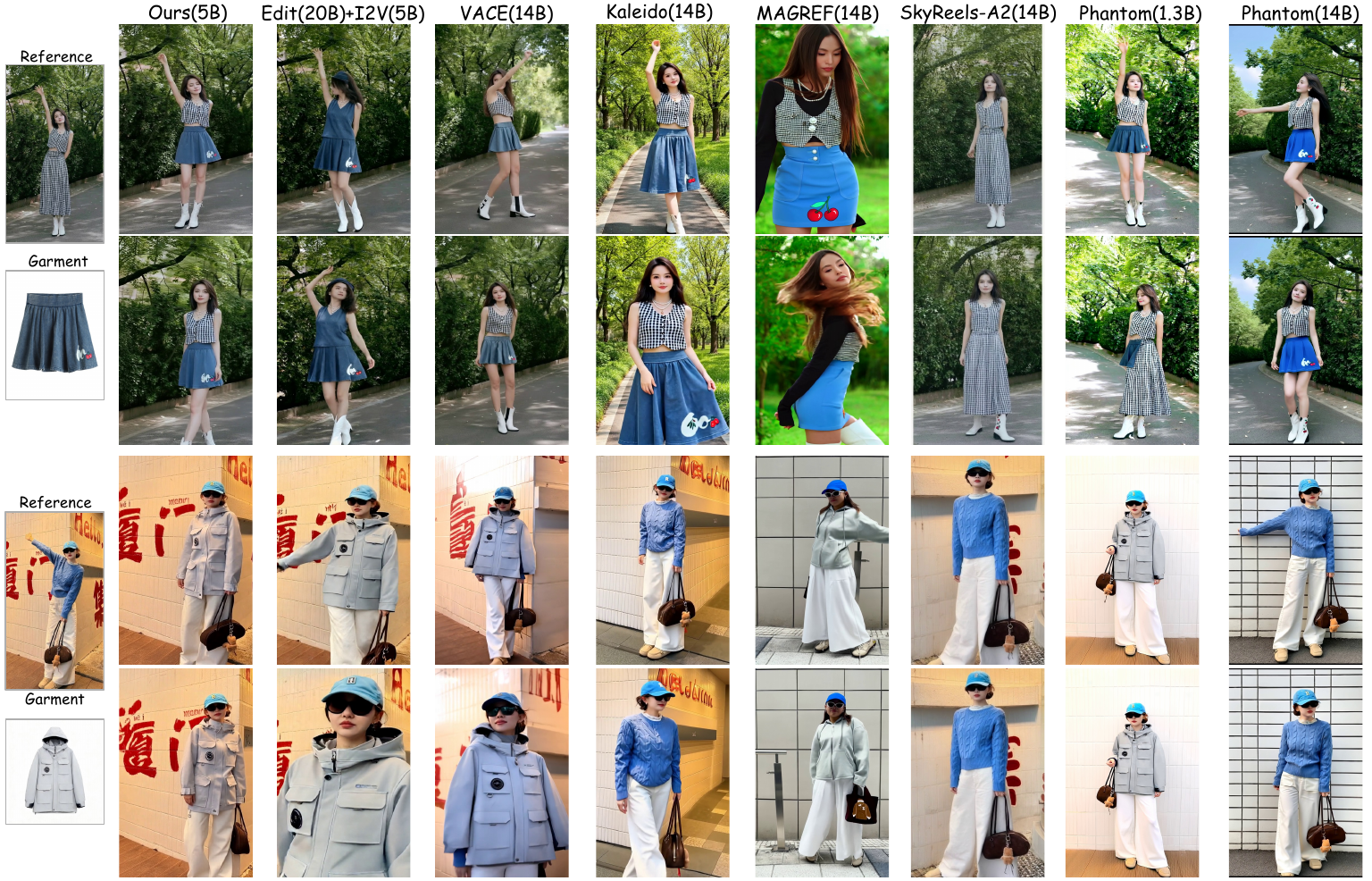}
   \caption{
   Additional qualitative comparison between our \method and other baselines.
   }
   \label{fig:additional}
\end{center}
\vspace{-1.0em}
\end{figure*}

\begin{figure*}[!h]
\begin{center}
   \includegraphics[width=0.98\textwidth] {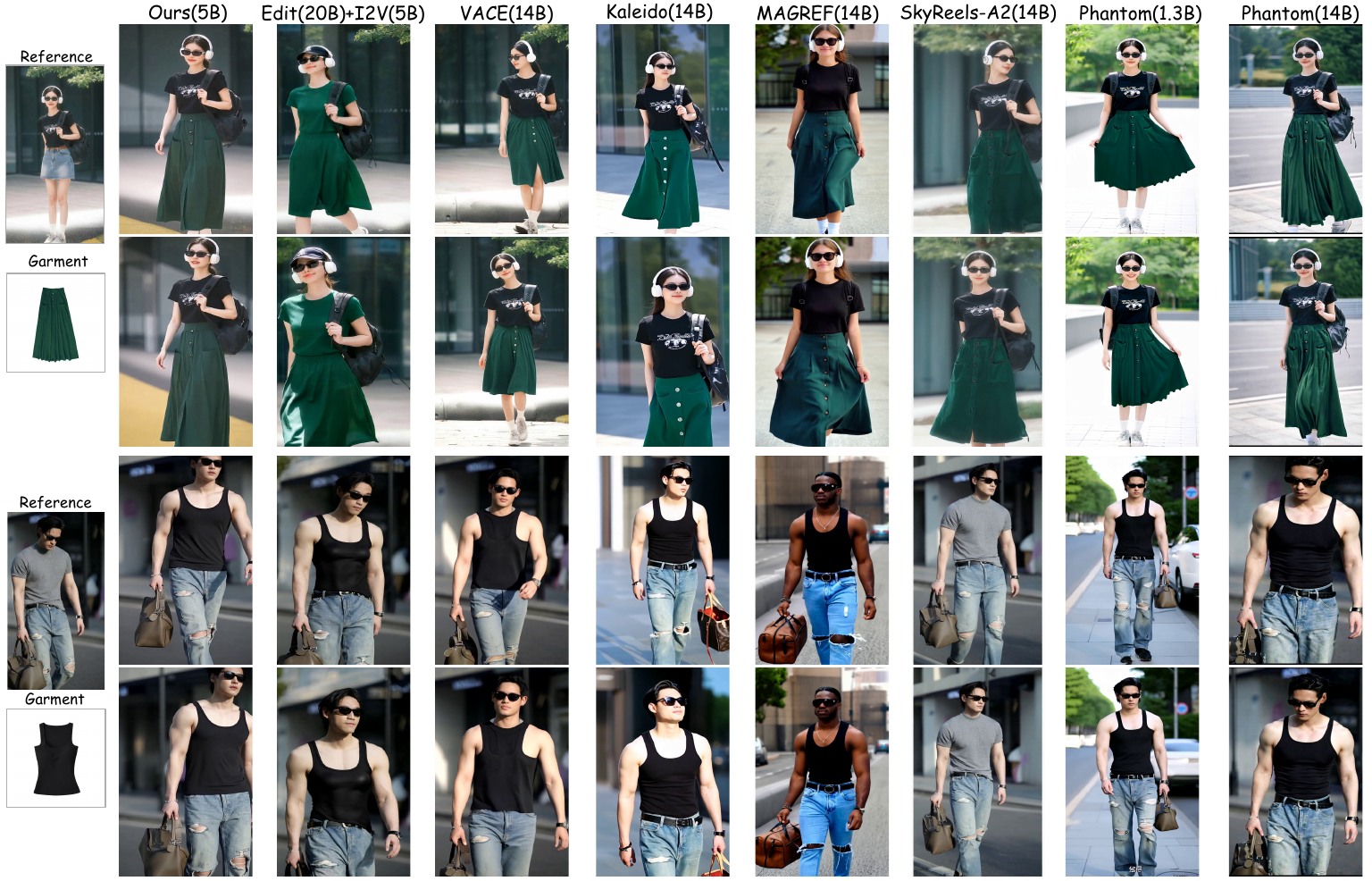}
   \caption{
   Additional qualitative comparison between our \method and other baselines.
   }
   \label{fig:additional2}
\end{center}
\vspace{-1.0em}
\end{figure*}

To further validate the effectiveness of  our \method and its advantages over competing baselines, we provide additional qualitative comparisons in Figure~\ref{fig:additional} and Figure~\ref{fig:additional2}.
Visually, \method demonstrates better character consistency and garment consistency, while producing more coherent and higher-quality results.

\section{Additional Examples of Short Video Customization}
\begin{figure*}[!h]
\begin{center}
   \includegraphics[width=0.98\textwidth] {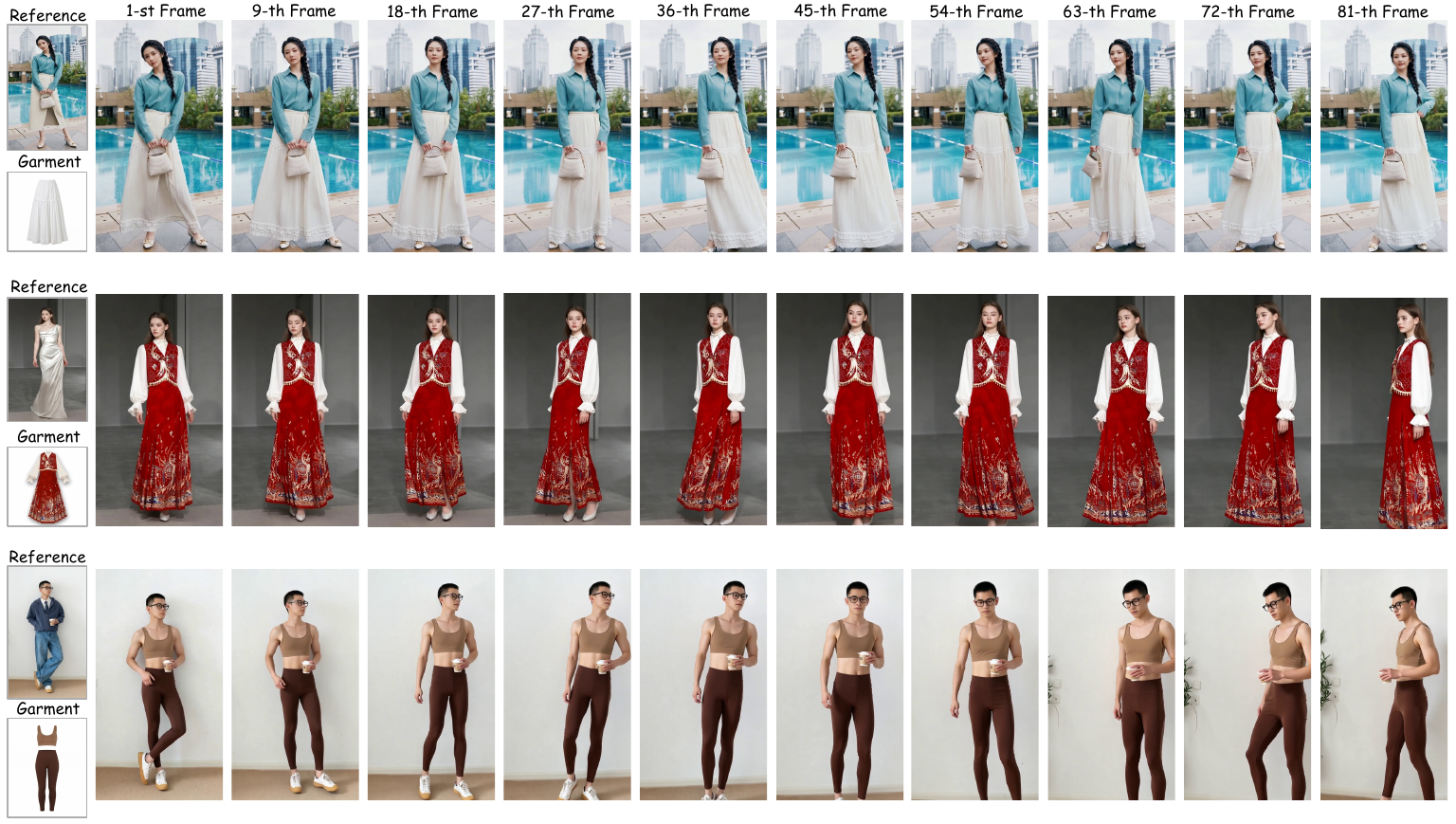}
   \caption{
   Additional results for short video customization using our \method.
   }
   \label{fig:short_video}
\end{center}
\end{figure*}

\begin{figure*}[!h]
\begin{center}
   \includegraphics[width=0.98\textwidth] {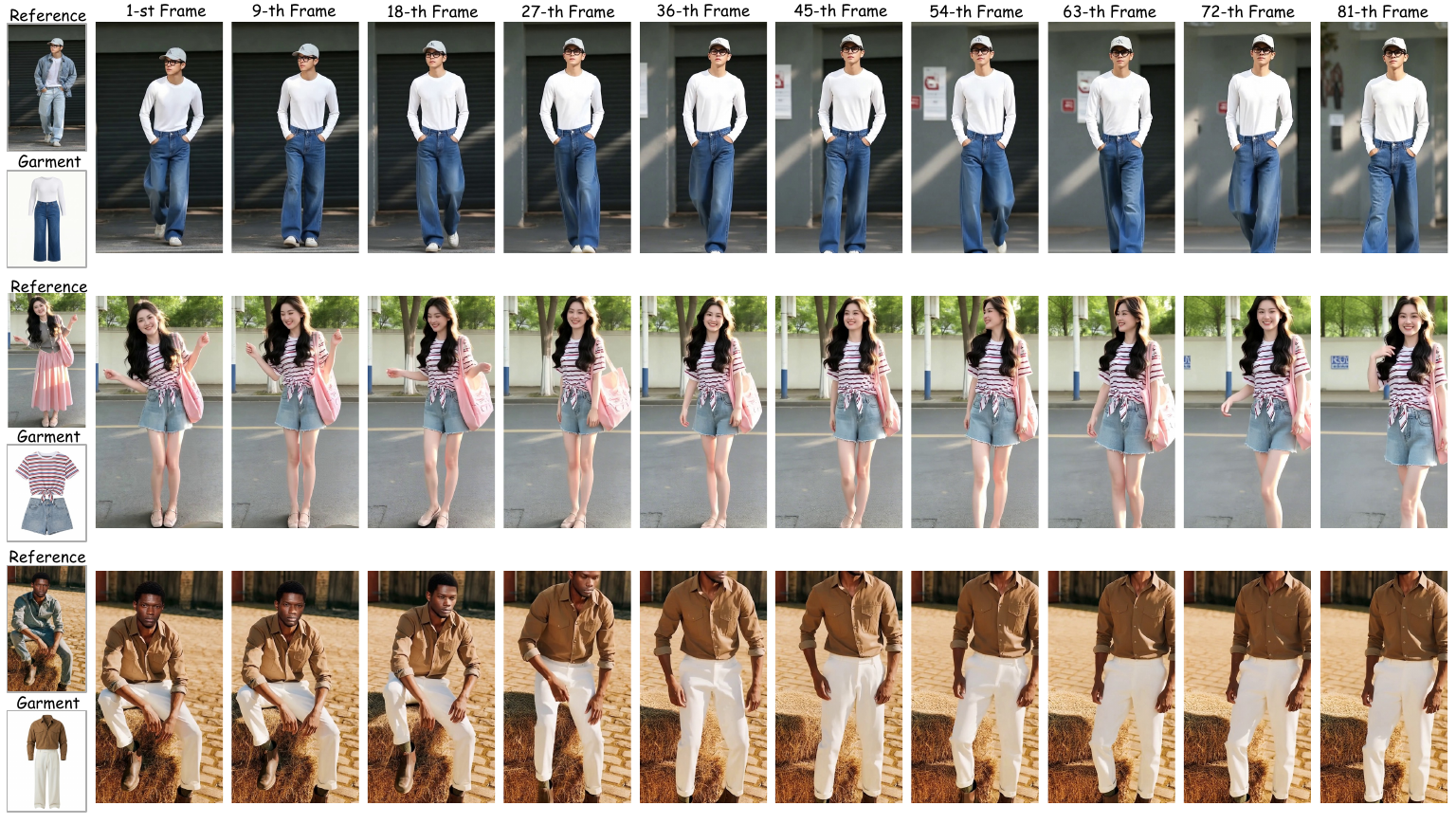}
   \caption{
   Additional results for short video customization using our \method.
   }
   \label{fig:short_video2}
\end{center}
\end{figure*}

Our \method is trained on 81-frame video clips and therefore supports customized generation of short videos of the same length. We provide additional examples, as shown in Figure~\ref{fig:short_video} and Figure~\ref{fig:short_video2}.
Notably, \method can produce coherent and high-fidelity human-garment customized videos, further highlighting its superiority.

\section{Additional Examples of Interactive Customization}
\begin{figure*}[!h]
\begin{center}
   \includegraphics[width=0.98\textwidth] {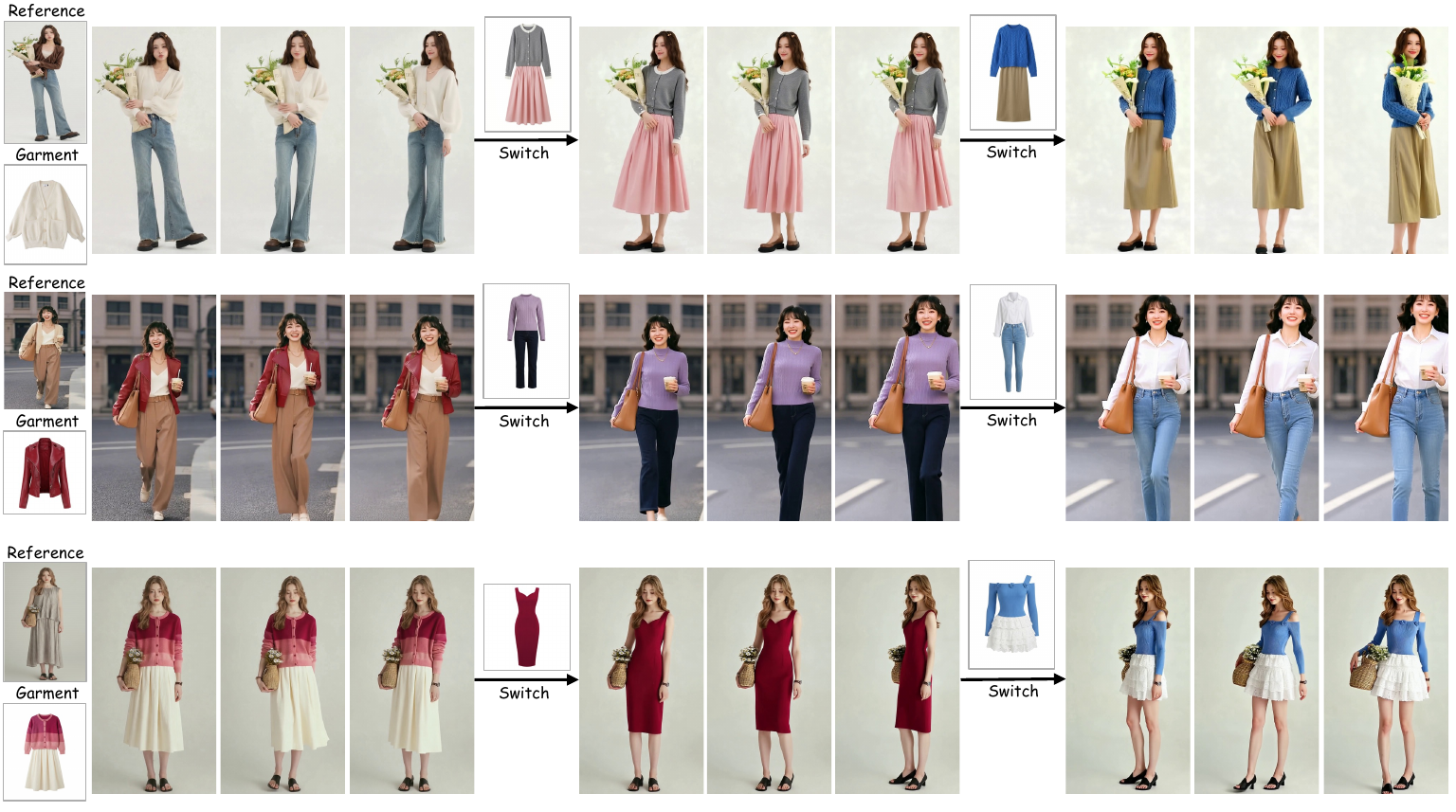}
   \caption{
   Additional visualizations for interactive multi-garment video customization using our \method.
   }
   \label{fig:switch}
\end{center}
\vspace{-1.0em}
\end{figure*}

\begin{figure*}[!h]
\begin{center}
   \includegraphics[width=0.98\textwidth] {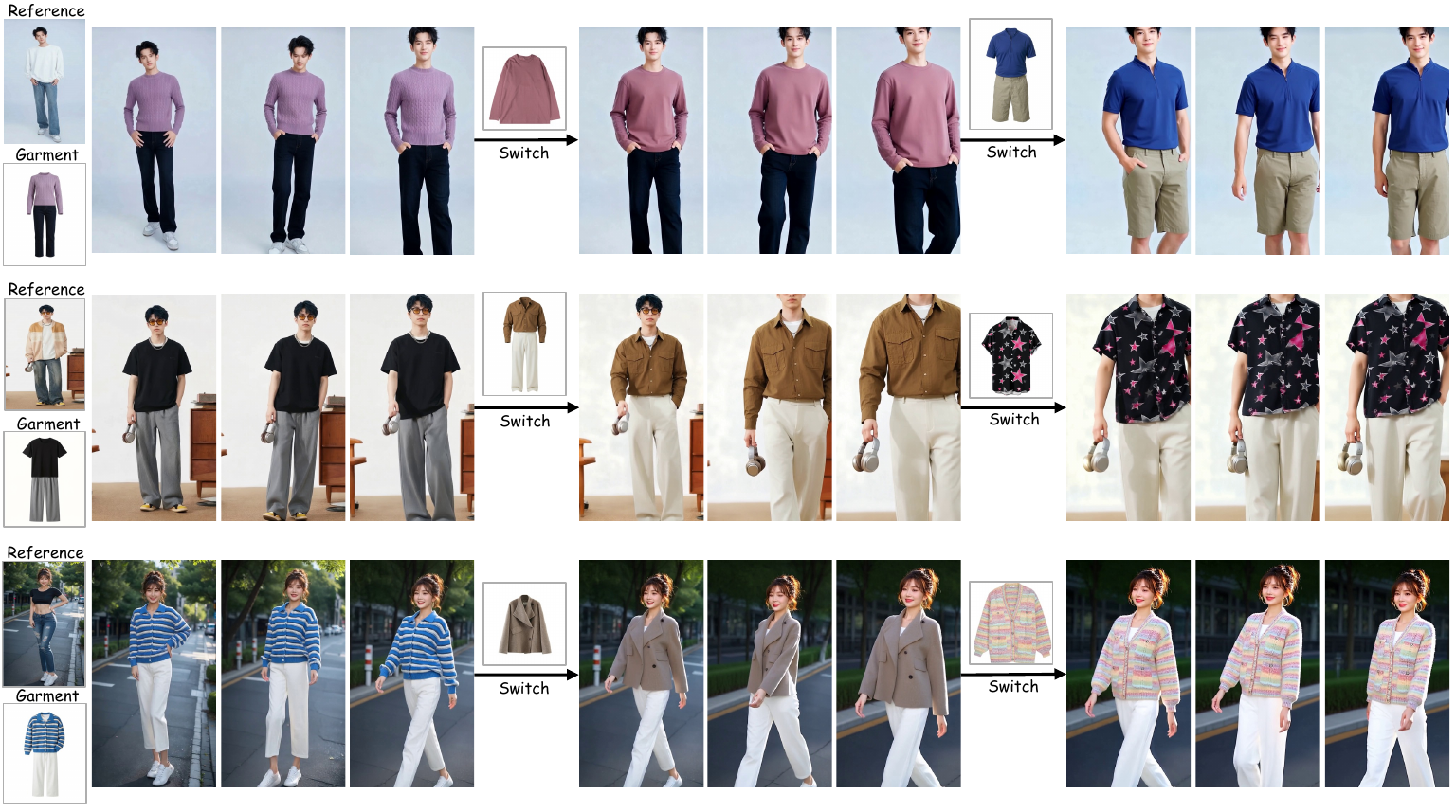}
   \caption{
   Additional visualizations for interactive multi-garment video customization using our \method.
   }
   \label{fig:switch2}
\end{center}
\vspace{-1.0em}
\end{figure*}

Thanks to the proposed KV cache rescheduling strategy, our \method supports interactive multi-garment customized generation, with the additional examples shown in Figure~\ref{fig:switch} and Figure~\ref{fig:switch2}.
Unlike conventional methods that require a reference image to be specified in advance, \method allows users to freely switch reference images at different stages of generation while preserving motion continuity, enabling interactive customization.
This further demonstrates the superiority of \method in the interactive generation domain.

\section{Additional Examples of Long Video Customization.}
\begin{figure*}[!h]
\begin{center}
   \includegraphics[width=0.98\textwidth] {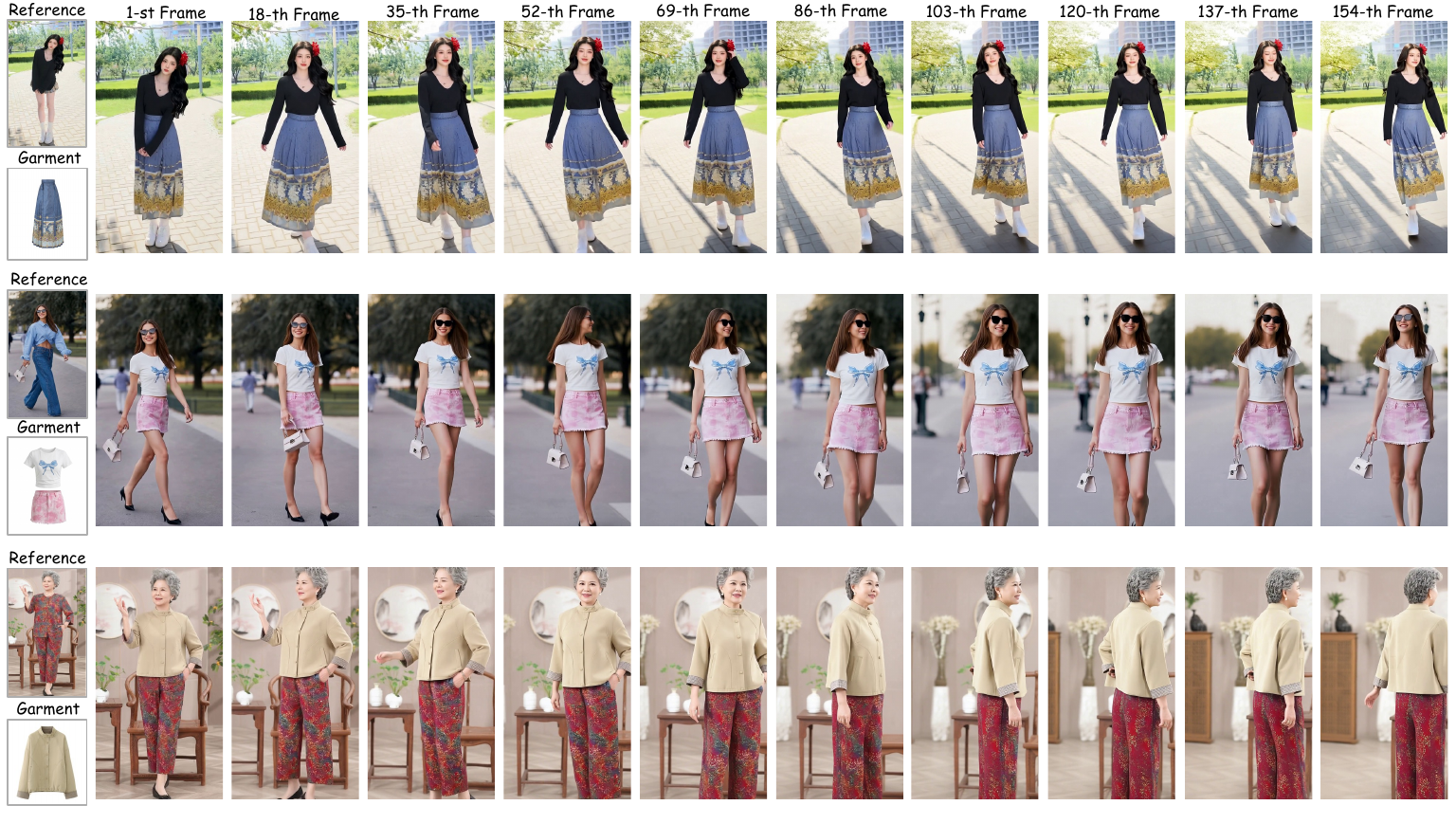}
   \caption{
   Additional long video extrapolation visualizations of our \method.
   }
   \label{fig:long_video}
\end{center}
\end{figure*}

\begin{figure*}[!h]
\begin{center}
   \includegraphics[width=0.98\textwidth] {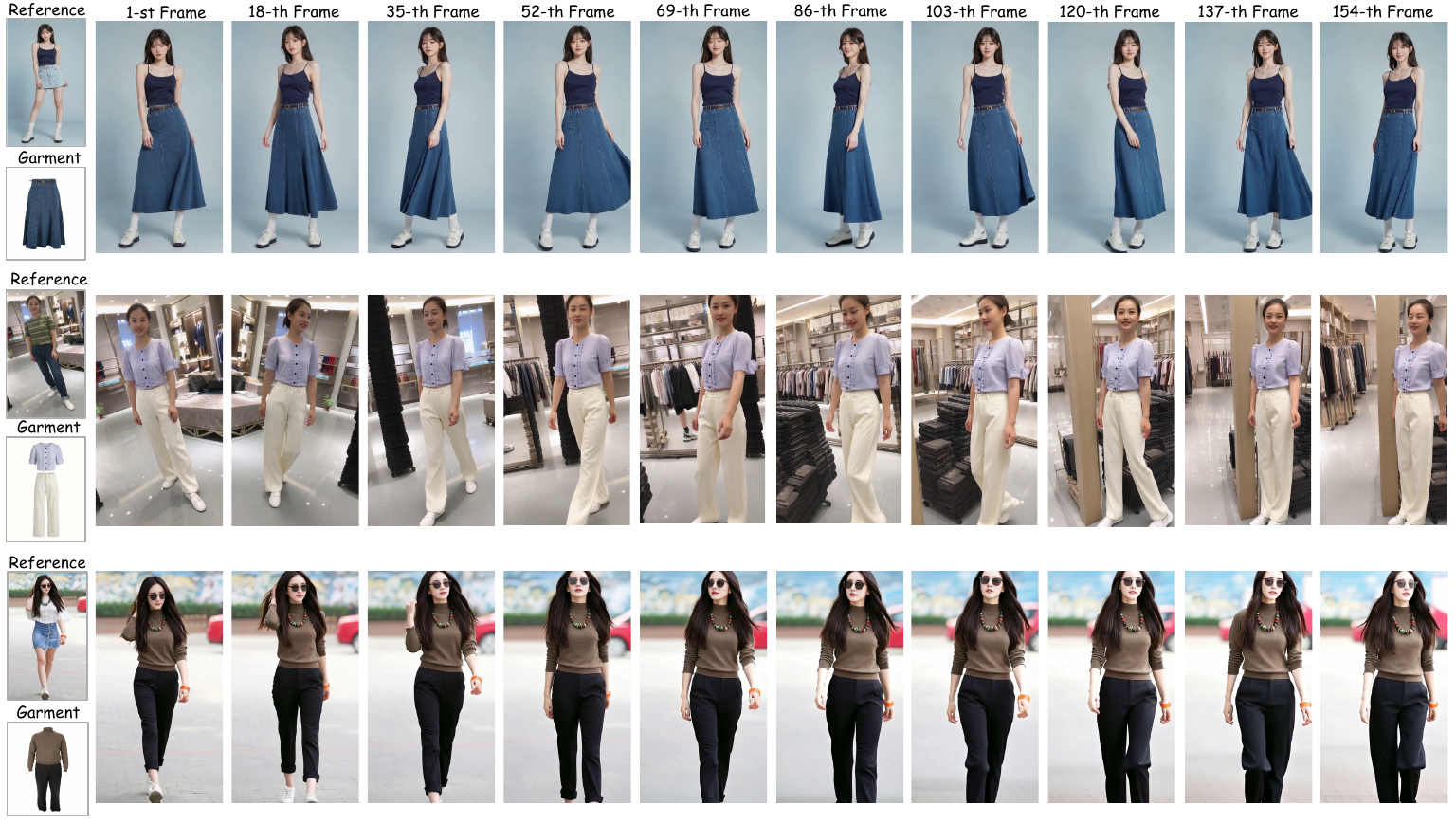}
   \caption{
   Additional long video extrapolation visualizations of our \method.
   }
   \label{fig:long_video2}
\end{center}
\end{figure*}

Benefiting from our dedicated autoregressive design, \method can generalize beyond the training sequence length, thereby enabling customized generation of longer videos. Additional qualitative results are provided in Figure~\ref{fig:long_video} and Figure~\ref{fig:long_video2}.
The qualitative results show that \method maintains long-range character consistency and garment consistency.

\section{Prompt List of Figures}
For reproducibility, we list the prompts used to generate Figure~\ref{fig:teaser} in the main paper:
\begin{enumerate}
\item ``A woman wearing a blue beret, earrings, and a watch stands on a floral garden path. She takes light steps forward, with her arms swinging naturally. Her gaze shifts from downward to focusing on the lens with a gentle smile, then smoothly transitions into a still pose, ensuring the movement is continuous and physically realistic.''
\end{enumerate}

For reproducibility, we list the prompts used to generate Figure~\ref{fig:analysis} in the main paper:
\begin{enumerate}
\item ``A woman performs a series of poses in an indoor setting while holding a white handbag in her right hand. Initially facing the camera, she subtly shifts her body to the left and places her left hand into her pocket. She then moves her left hand to rest lightly on a black shelving unit behind her. Throughout the video, she maintains a friendly smile and steady eye contact with the camera, with subtle changes in her stance and orientation. The video is filmed in a minimalist indoor studio featuring plain white walls and a light grey carpeted floor. To the right, a sleek black shelf displays decorative items such as vinyl records and magazines, while the corner of a white sofa is partially visible on the left. The lighting is bright and diffused, creating a clean and modern aesthetic. The camera remains stationary in a full-body composition, ensuring a consistent visual style.''
\end{enumerate}

For reproducibility, we list the prompts used to generate Figure~\ref{fig:qualitative_comparison} in the main paper:
\begin{enumerate}
\item ``A man strolls along an outdoor brick path, wearing a brown turtleneck long-sleeved knit sweater paired with white shorts and beige sandals. He maintains a steady forward gait, his arms swinging naturally to showcase the drape of the new garment. The camera performs a smooth tracking shot, moving backward to keep him centered in the frame. Initially looking to the side, he slowly turns his head forward, shifting his gaze naturally and smoothly to look directly into the lens.''
\item ``A young woman stands in a room, wearing a red short-sleeved t-shirt paired with a long floral skirt, with a red string bracelet on her left wrist. She initially tilts her head slightly to the side, then naturally shifts her gaze back to the lens with a soft smile. She performs a subtle turn to the left, causing the hem of the long skirt to sway with natural physics. The camera pans slowly to the right to keep her centered as she turns back to face forward, showcasing the elegant silhouette of the outfit.''
\end{enumerate}

For reproducibility, we list the prompts used to generate Figure~\ref{fig:app} in the main paper:
\begin{enumerate}
\item ``A young woman walks near park flowers, wearing a blue zippered crop top and lace-up distressed denim shorts, accented with a white cap, necklace, and a bag featuring a teddy bear charm. She walks forward with an elegant catwalk stride, her arms swinging naturally while her platform sneakers land steadily. The camera performs a steady tracking shot, keeping her centered. She shifts her gaze from forward to the lens, blinking with a smile and tilting her head slightly.''
\item ``A young woman stands against a pink and blue background. She wears purple flower earrings and carries a pink woven bag on her shoulder. She walks forward with light steps, her arms swinging naturally, while the bag strap bounces slightly. She then tilts her head toward the camera with a bright smile and a natural blink. The movement is smooth and consistent, ending in a frozen mid-stride pose.''
\item ``A young woman stands in a room filled with books and vintage items. She wears a baseball cap with text and has one hand in her pocket. She slowly lowers her hand from the cap, shifts her weight, and turns slightly to the right. Her gaze shifts from the lens toward the stack of books before turning back to blink and smile naturally. The movement is smooth and consistent, ending with her holding a slightly turned pose.''
\end{enumerate}

For reproducibility, we list the prompts used to generate Figure~\ref{fig:ablation2} in the main paper:
\begin{enumerate}
\item ``In the video, a young woman slowly enters from the right side of the frame and stops near a table. She initially looks down in reflection, then gracefully turns her head to the left, gazing into the distance. The camera remains in a fixed position, capturing the scene through a transparent glass door, with subtle reflections on the glass shifting as she moves. The setting is an interior space with soft lighting, likely a cafe or restaurant, featuring wooden tables and chairs with a warm texture. The overall visual style is realistic and cinematic, using the glass door in the foreground to create an observational perspective within a warm and tranquil atmosphere.''
\item ``Captured from a static camera angle, a young woman with long, flowing black hair sways her body gracefully to a rhythmic beat. She raises her left hand to touch and adjust her hair, tossing it over her shoulder while her arms move naturally in sync with her shifting posture. Throughout the sequence, she maintains direct eye contact with the camera, exhibiting a series of fluid and confident movements. The setting is a minimalist and elegant indoor environment featuring large beige pleated curtains in the background and a brown striped carpet on the floor. To the right stands a contemporary white floor lamp with a decorative stem made of transparent spherical crystals. The lighting is soft and diffused, dominated by a warm color palette of beige and tan, creating a cozy, high-quality lifestyle aesthetic.''
\end{enumerate}

For reproducibility, we list the prompts used to generate Figure~\ref{fig:additional} and Figure~\ref{fig:additional2} in the Appendix:
\begin{enumerate}
\item ``On a lush tree-lined path, a woman wears a black and white checkered vest paired with a blue mini skirt featuring a cherry graphic, accented by a pearl necklace and white boots. She slowly lowers her raised right arm and turns her body slightly to the left to showcase the skirt's silhouette. The camera orbits steadily around her in an arc. She shifts her gaze from the side back to the lens, her long hair swaying naturally over her shoulders as she moves.''
\item ``A woman in a blue cap and sunglasses stands by a white tiled wall, wearing a light grey multi-pocket hooded jacket, white wide-leg pants, and beige shoes, holding a brown bag with a bear charm. She slowly lowers her raised right arm and takes a natural step forward to showcase the outfit. The camera pans horizontally to the right; she turns her head from the side to face forward, gazing into the lens through her sunglasses with a relaxed posture.''
\item ``A young woman stands outdoors wearing white headphones and sunglasses, dressed in a black short-sleeved T-shirt and a dark green button-front maxi skirt, carrying a black backpack with white socks and sneakers. She walks steadily toward the camera, the long skirt's hem swaying naturally and gracefully with her steps. The camera pulls back smoothly to reveal the full silhouette of the outfit; she shifts her gaze from the side to the lens, smiling faintly and blinking.''
\item ``On a city street, a black-haired man wearing sunglasses is dressed in a black U-neck tank top paired with ripped blue jeans and a black belt, holding a brown leather bag in his right hand with a watch and bracelet on his wrists. He walks forward with steady steps, his body swaying naturally to showcase the fit of the tank top. The camera slowly zooms out from a close-up to a full-body view. He shifts his gaze from downward to looking straight ahead with a calm expression.''
\end{enumerate}

For reproducibility, we list the prompts used to generate Figure~\ref{fig:short_video} and Figure~\ref{fig:short_video2} in the Appendix:
\begin{enumerate}
\item ``A young woman stands by the poolside with city buildings in the background, wearing a turquoise long-sleeved shirt and a white tiered ruffled long skirt, holding a small cream-colored handbag. She walks toward the camera with light catwalk steps, the layered hem swaying naturally. The camera slowly zooms out to reveal the full silhouette; her gaze shifts from the side back to the lens as she gives a slight, steady nod.''
\item ``A young woman stands in a minimalist gray indoor setting, wearing a white puff-sleeved blouse and a red phoenix-embroidered vest paired with a red patterned pleated skirt, with a thin bracelet on her left wrist. She looks down initially, then raises her gaze to the camera while turning slightly to the left, allowing the skirt to drape naturally. The camera smoothly pulls back from a close-up to reveal the full-length silhouette of the traditional outfit.''
\item ``In front of a white wall, a man wearing black-rimmed glasses holds a coffee cup, dressed in a tan sports bra and dark brown leggings with white sneakers. He slowly transitions from a leaning pose to a steady upright stance, balancing his weight on both feet to showcase the silhouette. The camera zooms out smoothly to capture the full outfit; he tilts his head slightly, shifting his gaze from the side back to the lens with a calm expression.''
\item ``A young man wearing a baseball cap and black glasses stands before a dark rolling shutter, dressed in a white long-sleeved top and dark blue wide-leg trousers. He moves his hands out of his pockets to his sides and walks forward toward the camera, the loose pant legs creating natural folds and swaying with each step. The camera tracks him steadily; he tilts his head slightly upward, shifting his gaze from the side to the lens with a composed expression.''
\item ``A young woman stands by an outdoor road, wearing a red and blue striped tie-front top with light-blue denim shorts and carrying a large pink canvas bag. Transitioning from an open-arm pose, she naturally lowers her hands and walks forward toward the camera with a brisk, steady gait. The camera tracks backward smoothly, keeping her centered in the frame. She briefly looks down before raising her head, shifting her gaze from the side to the lens with a bright smile, her long hair swaying naturally as she moves.''
\item ``A Black man sits on an outdoor hay bale, wearing a brown long-sleeved shirt with double chest pockets, paired with wide-leg white trousers, brown boots, and olive socks. Resting his hands on his knees, he slowly stands up from the bale, smoothing the shirt front to showcase the drape. The camera pulls back slowly to reveal the full outfit. He tilts his head slightly, shifting his gaze from the side back to the lens with a calm expression.''
\end{enumerate}

For reproducibility, we list the prompts used to generate Figure~\ref{fig:switch} and Figure~\ref{fig:switch2} in the Appendix:
\begin{enumerate}
\item ``The woman stands against a white backdrop holding an exquisite bouquet of lilies and greenery. Starting with a direct gaze, she blinks and transitions into a natural smile with gentle eyes. She then turns her body slowly to the right while holding the bouquet, showcasing her side profile with smooth movements. Her hair and the flower petals sway slightly following the physics of the motion. Finally, she holds a graceful side-facing posture with a relaxed expression.''
\item ``A woman strolls through an urban street. She carries a brown leather tote bag on her right shoulder and holds an iced coffee in her left hand, with her gold necklace and hair clip glinting. She walks forward toward the camera with light steps, her arms swinging naturally and the bag swaying slightly with her rhythm. Initially laughing and looking aside, she then turns her gaze to the camera with bright eyes, eventually pausing while maintaining a natural walking posture.''
\item ``A woman stands against a simple background, cradling a woven basket of white daisies in her right arm and wearing a watch on her left wrist. She initially looks down at the flowers, then slowly turns her body to the left with smooth movements, her arms swinging naturally. She then shifts her gaze to the camera with a gentle smile and a slight head tilt, ensuring a fluid transition before returning to a stable forward-facing pose.''
\item ``A young man stands against a clean light blue background. He shifts his center of gravity and takes a natural small step forward, with his arms swinging slightly and naturally. He blinks and tilts his head down slightly before looking up to gaze at the lens with a confident and gentle smile. His head turns slightly in coordination with his body, and the entire movement is smooth, consistent, and physically natural.''
\item ``A young man stands in a minimalist studio with a wooden cabinet nearby, holding a pair of headphones in his right hand. Wearing orange sunglasses and a silver chain, he begins by taking a steady step forward. As his weight shifts, he transitions from a slight head tilt to looking directly into the lens with a relaxed expression. The headphones sway gently with his movement, which is smooth and physically natural, ending in a stable standing pose.''
\item ``The woman stands in the center of a leafy street, wearing hoop earrings. She tilts her head slightly to showcase her accessories, then begins walking slowly toward the camera with her arms swinging naturally and her weight shifting steadily. During the walk, she turns her gaze from the side back to the lens, blinking naturally with a confident smile, before coming to a smooth stop.''
\end{enumerate}

For reproducibility, we list the prompts used to generate Figure~\ref{fig:long_video} and Figure~\ref{fig:long_video2} in the Appendix:
\begin{enumerate}
\item ``On a sunlit park path, a long-haired woman with a red flower hair accessory wears a black V-neck sweater paired with a long blue traditional skirt featuring gold patterns and a delicate necklace. She takes elegant catwalk steps toward the camera, the heavy blue hem swaying naturally with her stride. The camera moves backward smoothly to track her, maintaining a consistent frame. She tilts her head slightly, shifting her gaze upward from the ground to fixate on the lens with a gentle smile.''
\item ``On an outdoor park path, a long-haired woman wearing sunglasses is dressed in a white short-sleeved T-shirt with a blue bow and a pink tie-dye denim mini skirt. She carries a white mini handbag in her left hand and holds a phone in her right. She walks toward the camera with a graceful catwalk gait, her movements fluid and natural. The camera performs a steady tracking shot as she tilts her head slightly to the right, shifting her gaze from the side back to the lens with a smile, her hair swaying gently with her steps.''
\item ``A silver-haired elderly woman stands by a traditional wooden chair, wearing a beige stand-collar jacket with plaid cuffs and a cinched hem, paired with red printed trousers and a pearl necklace. She lowers her raised right arm and gently turns to the left to display the jacket's side profile. The camera pulls back steadily to capture the full ensemble; the woman turns her head to shift her gaze from the side back to the lens with a kind and composed expression.''
\item ``A young woman stands against a light blue background, wearing a navy blue camisole paired with a long dark blue denim skirt featuring a brown belt, along with white socks and sneakers. She slowly turns her body to the left, showcasing the drape of the long skirt and the belt details with fluid movements. The camera performs a subtle orbital rotation around her; she tilts her chin slightly and shifts her gaze naturally from the side back to the lens with a gentle smile.''
\item ``A young woman stands in a clothing store wearing a light purple ruffled short-sleeve shirt and cream-colored wide-leg pants, with a gold bracelet on her right wrist and white sneakers. She walks toward the camera with light steps, the wide pant legs swaying naturally with her movement. The camera tracks her steadily; she initially looks toward the side shelves before gently turning her head to shift her gaze back to the lens with a smile.''
\item ``A long-haired woman wearing sunglasses, a colorful necklace, and an orange bracelet, dressed in a brown turtleneck sweater and black trousers with white sneakers, walks outdoors. She maintains a steady gait approaching the camera, her arms swinging naturally to showcase the drape of the sweater and trousers. The camera tracks her movement, keeping her centered in the frame. She tilts her head slightly to the left, shifting her gaze from the side back to the lens with a relaxed expression.''
\end{enumerate}

\section{System Prompts of VLM}
\label{sec:system_prompt}
We present the system prompt for Gemini-3.1 to generate prompts in training datasets below:
\begin{tcolorbox}[
    colback=gray!10,
    colframe=gray!30,
    boxrule=0.3pt,
    arc=1.5pt,
    left=6pt,
    right=6pt,
    top=6pt,
    bottom=6pt
]
\textbf{System Prompt:}
\\
Please generate a structured multilingual description based on the input video content, strictly following the specifications below:
\\
1. \textit{Dynamic element description}: Focus on content that changes over time in the video, such as: (a) The subject (e.g., person, animal, object, etc.) and changes in its behavior, state, or position; (b) Specific actions (e.g., running, waving, opening a door, vehicles moving, etc.); (c) Scene transitions (e.g., switching from a street to an indoor setting, weather changing from sunny to rainy, etc.); (d) Camera movement (e.g., push-in, pull-out, pan, tracking shot, fixed shot, zoom, etc.).
\\
2. \textit{Static element description}: Accurately identify visual features that remain unchanged throughout the video, such as: (a) The inherent appearance of the subject or environment (e.g., clothing style, architectural style, object color and material, indoor layout, etc.); b The overall aesthetic style (e.g., realistic, animated, film-like, cyberpunk, minimalist, etc.); (c) Consistent visual style elements such as color tone, lighting atmosphere, and composition principles; d. The character’s clothing, styling, and accessories.
\\
3. \textit{Additional notes:} In the dynamic element description, completely ignore any description of the character’s clothing, styling, or accessories, but handheld items (if any) may be briefly described.
\\
Please output in standard JSON format, containing the following four fields. The value of each field must be an array of two strings: (a) The first string: description of dynamic elements; (b) The second string: description of static elements.
\\
The field definitions are as follows:
\\
``cn long'': two detailed Chinese descriptions (the first for dynamics, the second for statics);
\\
``cn short'': two concise Chinese descriptions (the first for dynamics, the second for statics);
\\
``en long'': two detailed English descriptions, semantically corresponding to cn long;
\\
``en short': two concise English descriptions, semantically corresponding to cn short.
\\
\textit{Output requirements}:
\\
1. Use natural, objective, and accurate language, based only on visible video content, without adding speculation or fabricated details;
\\
2. The Chinese and English descriptions should be semantically aligned, but do not need to be word-for-word translations;
\\
3. Long descriptions should be comprehensive and detailed, while short descriptions should be concise and focused on the core information.
\end{tcolorbox}

We present the system prompt for Gemini-3.0 to generate prompts in HGC-Bench below:
\begin{tcolorbox}[
    colback=gray!10,
    colframe=gray!30,
    boxrule=0.3pt,
    arc=1.5pt,
    left=6pt,
    right=6pt,
    top=6pt,
    bottom=6pt
]
\textbf{System Prompt:}
\\
Role Definition:
\\
You are a senior prompt expert specializing in video generation (I2V). Your core task is to write coherent, dynamic, and physically plausible video-generation description scripts based on the characteristics of the first-frame image [Image1] and the target garment image [Image2]. The script mainly includes a static description of the first frame and a subsequent dynamic description. The static description should focus on the person wearing the new garment in the original scene. The dynamic description should be coherent and natural, avoiding motion collapse.
\\
I. \textbf{Static Description}:
\\
    1. \textit{Scene Description}: Since the first frame largely provides the scene information, only a brief description is needed here.
    2. \textit{Garment Description}: The person’s original clothing in the first-frame image [Image1] must be used as an anchor, with its type described but without detailed description, and integrated with the new garment from the target garment image [Image2], whose type should be described but without unnecessary details. In the description, directly assign the new garment to the person. It is strictly forbidden to use words that describe a dynamic transformation process, such as ``changed into,'' ``switched to,'' ``turned into,'' or similar expressions. The new garment should already be part of the person’s outfit in the initial state.
    3. \textit{Detail Description}: Pay attention to describing accessories, backpacks, handheld items, and similar details of the person in the first-frame image [Image1]. Ignoring these details may degrade the performance of the I2V task. The description should focus on the person wearing the new garment in the original scene.
    4. \textit{Consistency Preservation}: When the target garment image [Image2] provides only part of the outfit, the description must logically and coherently match it with the remaining clothing from [Image1], ensuring overall character consistency. Do not describe the old garment in the first-frame image [Image1] that has been replaced, because it has already been discarded and replaced by the new garment.
    5. \textit{Hallucination Avoidance}: It is strictly forbidden to use any adjectives describing visual style, such as ``cinematic,'' ``high-definition,'' or ``hyper-realistic.'' It is strictly forbidden to fabricate objects that do not exist in the first-frame image.
\\
II. \textbf{Dynamic Description (one item must be selected from each category)}:
\\
    1. \textit{Overall Body Movement}: One of the following must be randomly included:
        (a) Runway Walk / Walking: Walking forward toward the camera, walking away with the back facing the camera, catwalk, etc. Reasonable imagination is allowed, but the motion must be coherent and should avoid causing collapse.
        (b) Turning / Spinning: Slightly turning to the left, turning backward, etc. Reasonable imagination is allowed, but the motion must be coherent and should avoid causing collapse.
        (c) Posture Transition: Standing up from a seated pose, sitting down from a standing pose, shifting from leaning to upright, etc. Reasonable imagination is allowed, but the motion must be coherent and should avoid causing collapse.
        (d) Stretching / Extending: Raising both arms horizontally, stretching upward, turning sideways to show the back or side cut, etc. Reasonable imagination is allowed, but the motion must be coherent and should avoid causing collapse.
\\
    2. \textit{Facial Expression and Head Movement}: One of the following must be randomly included:
        (a) Gaze Shift: From looking down to looking at the camera, glancing sideways and then turning back, blinking with a slight smile, etc. Reasonable imagination is allowed, but the motion must be coherent and should avoid causing collapse.
        (b) Head Movement: Tilting the head left and right for display, hair-swaying motion, etc. Reasonable imagination is allowed, but the motion must be coherent and should avoid causing collapse.
        (c) Notes:
            (i) Motion Stability: The description must logically include an “initial transition phase,” a “dynamic display phase,” and a “final freeze phase,” but these do not need to be explicitly written out. Reasonable imagination is allowed, but the motion must be coherent and should avoid causing collapse.
            (ii) Physical Plausibility: The range of motion must be kept within a reasonable scope. For example, turning should preferably not exceed 90 degrees to avoid limb deformation. The motion must follow gravity, such as the hem of the garment moving with the body during turning and the arms swinging naturally.
            (iii) First-Frame Continuity: If there is a large difference between the pose in the first frame and the subsequent pose, the description should provide a gradual, coherent, and smooth transition to avoid collapse.
            (iv) Hallucination Avoidance: It is strictly forbidden to use any adjectives describing visual style, such as ``cinematic,'' ``high-definition,'' or ``hyper-realistic.''
\\
III. \textbf{Output Requirements}:
\\
1. Each set of descriptions must be strictly limited to within 150 words.
\\
2. Output format: JSON, generating both Chinese (cn short) and English (en short) descriptions.
\end{tcolorbox}

We present the system prompt for Gemini-3.0 to evaluate garment consistency below:
\begin{tcolorbox}[
    colback=gray!10,
    colframe=gray!30,
    boxrule=0.3pt,
    arc=1.5pt,
    left=6pt,
    right=6pt,
    top=6pt,
    bottom=6pt
]
\textbf{System Prompt:}
\\
Task Objective: Evaluate the quality of AI-customized video generation.
\\
Input Description:
[Image 1]: Original image of the model (reference for person identity, pose, non-target clothing, and original background).
[Image 2]: Target garment image (reference for the clothing to be virtually tried on).
[Video 1]: AI-generated video sequence (evaluation target).
\\
Scoring Dimensions: 1–5, where 1 is the worst and 5 is the best:
\\
\textit{1. High-level garment consistency}
\\
Evaluate how well the target garment in [Image 2] matches the garment worn by the model in the corresponding [Video 1] sequence at a high-level semantic level.
Checkpoints: whether the category, overall silhouette (e.g. fit, cut), large color block distribution, and overall style are consistent.
Execution focus: Only evaluate the match of the target garment from [Image 2]; ignore other non-target garments worn by the model in [Video 1].
\\
\textit{2. Low-level garment consistency}
\\
Evaluate how well the target garment in [Image 2] matches the garment worn by the model in the corresponding [Video 1] sequence at the pixel-detail level.
Checkpoints: whether fine-grained features are accurately reproduced, such as patterns (prints, stripes), logos, embroidery, fabric texture, surface gloss, etc.
Hard constraint: If there is obvious pattern distortion, blurred logos, or completely incorrect fabric texture, this score must be <= 2.
Execution focus: Only evaluate the match of the target garment from [Image 2]; ignore other non-target garments worn by the model in [Video 1].
\\
\textit{3. Non-target garment preservation}
\\
Evaluate the consistency between the non-target garments in [Video 1] (i.e. all other clothing items, accessories, etc. worn by the model besides the target garment) and those in [Image 1].
Checkpoints: whether the style, color, and texture of all non-target parts in [Video 1] have been incorrectly modified or removed.
Logical focus: Apart from virtually trying on the target garment, all other garments on the model in [Video 1] should preserve the original appearance from [Image 1] as much as possible.
\\
\textit{Output Requirements}:
\\
Please output only a single JSON object, without any explanatory text:
\\
\{
``high-level garment consistency'': 0-5,
``low-level garment consistency'': 0-5,
``non-target garment preservation'': 0-5
\}
\end{tcolorbox}

%% file: reference.bib
@String(PAMI  = {IEEE Transactions on Pattern Analysis and Machine Intelligence})

@String(CVPR  = {IEEE Conference on Computer Vision and Pattern Recognition})

@String(ICCV  = {International Conference on Computer Vision})

@String(ECCV  = {European Conference on Computer Vision})

@String(NeurIPS = {Conference on Neural Information Processing Systems})

@String(ICLR  = {International Conference on Learning Representations})

@String(TMM   = {IEEE Transactions on Multimedia})

@String(ACMMM = {ACM International Conference on Multimedia})

@String(PR    = {Pattern Recognition})

@String(SIGGRAPH = {ACM SIGGRAPH Conference on Computer Graphics and Interactive Techniques})

@article{yang2024cogvideox,
  title={Cogvideox: Text-to-video diffusion models with an expert transformer},
  author={Yang, Zhuoyi and Teng, Jiayan and Zheng, Wendi and Ding, Ming and Huang, Shiyu and Xu, Jiazheng and Yang, Yuanming and Hong, Wenyi and Zhang, Xiaohan and Feng, Guanyu and others},
  journal={arXiv preprint arXiv:2408.06072},
  year={2024}
}

@article{kong2024hunyuanvideo,
  title={Hunyuanvideo: A systematic framework for large video generative models},
  author={Kong, Weijie and Tian, Qi and Zhang, Zijian and Min, Rox and Dai, Zuozhuo and Zhou, Jin and Xiong, Jiangfeng and Li, Xin and Wu, Bo and Zhang, Jianwei and others},
  journal={arXiv preprint arXiv:2412.03603},
  year={2024}
}

@article{wan2025wan,
  title={Wan: Open and advanced large-scale video generative models},
  author={Wan, Team and Wang, Ang and Ai, Baole and Wen, Bin and Mao, Chaojie and Xie, Chen-Wei and Chen, Di and Yu, Feiwu and Zhao, Haiming and Yang, Jianxiao and others},
  journal={arXiv preprint arXiv:2503.20314},
  year={2025}
}

@article{sun2024autoregressive,
  title={Autoregressive model beats diffusion: Llama for scalable image generation},
  author={Sun, Peize and Jiang, Yi and Chen, Shoufa and Zhang, Shilong and Peng, Bingyue and Luo, Ping and Yuan, Zehuan},
  journal={arXiv preprint arXiv:2406.06525},
  year={2024}
}

@article{kondratyuk2023videopoet,
  title={Videopoet: A large language model for zero-shot video generation},
  author={Kondratyuk, Dan and Yu, Lijun and Gu, Xiuye and Lezama, Jos{\'e} and Huang, Jonathan and Schindler, Grant and Hornung, Rachel and Birodkar, Vighnesh and Yan, Jimmy and Chiu, Ming-Chang and others},
  journal={arXiv preprint arXiv:2312.14125},
  year={2023}
}

@article{lipman2022flow,
  title={Flow matching for generative modeling},
  author={Lipman, Yaron and Chen, Ricky TQ and Ben-Hamu, Heli and Nickel, Maximilian and Le, Matt},
  journal={arXiv preprint arXiv:2210.02747},
  year={2022}
}

@inproceedings{yin2024improved,
  title={Improved distribution matching distillation for fast image synthesis},
  author={Yin, Tianwei and Gharbi, Micha{\"e}l and Park, Taesung and Zhang, Richard and Shechtman, Eli and Durand, Fredo and Freeman, William T},
  booktitle=NeurIPS,
  year={2024}
}

@inproceedings{yin2025slow,
  title={From slow bidirectional to fast autoregressive video diffusion models},
  author={Yin, Tianwei and Zhang, Qiang and Zhang, Richard and Freeman, William T and Durand, Fredo and Shechtman, Eli and Huang, Xun},
  booktitle=CVPR,
  year={2025}
}

@inproceedings{chen2024diffusion,
  title={Diffusion forcing: Next-token prediction meets full-sequence diffusion},
  author={Chen, Boyuan and Mart{\'\i} Mons{\'o}, Diego and Du, Yilun and Simchowitz, Max and Tedrake, Russ and Sitzmann, Vincent},
  booktitle=NeurIPS,
  year={2024}
}

@article{huang2025vid2world,
  title={Vid2world: Crafting video diffusion models to interactive world models},
  author={Huang, Siqiao and Wu, Jialong and Zhou, Qixing and Miao, Shangchen and Long, Mingsheng},
  journal={arXiv preprint arXiv:2505.14357},
  year={2025}
}

@article{sun2025worldplay,
  title={Worldplay: Towards long-term geometric consistency for real-time interactive world modeling},
  author={Sun, Wenqiang and Zhang, Haiyu and Wang, Haoyuan and Wu, Junta and Wang, Zehan and Wang, Zhenwei and Wang, Yunhong and Zhang, Jun and Wang, Tengfei and Guo, Chunchao},
  journal={arXiv preprint arXiv:2512.14614},
  year={2025}
}

@article{zhang2025matrix,
  title={Matrix-game: Interactive world foundation model},
  author={Zhang, Yifan and Peng, Chunli and Wang, Boyang and Wang, Puyi and Zhu, Qingcheng and Kang, Fei and Jiang, Biao and Gao, Zedong and Li, Eric and Liu, Yang and others},
  journal={arXiv preprint arXiv:2506.18701},
  year={2025}
}

@article{wu2025qwen,
  title={Qwen-image technical report},
  author={Wu, Chenfei and Li, Jiahao and Zhou, Jingren and Lin, Junyang and Gao, Kaiyuan and Yan, Kun and Yin, Sheng-ming and Bai, Shuai and Xu, Xiao and Chen, Yilei and others},
  journal={arXiv preprint arXiv:2508.02324},
  year={2025}
}

@article{mao2025yume,
  title={Yume-1.5: A Text-Controlled Interactive World Generation Model},
  author={Mao, Xiaofeng and Li, Zhen and Li, Chuanhao and Xu, Xiaojie and Ying, Kaining and He, Tong and Pang, Jiangmiao and Qiao, Yu and Zhang, Kaipeng},
  journal={arXiv preprint arXiv:2512.22096},
  year={2025}
}

@article{yesiltepe2025infinity,
  title={Infinity-rope: Action-controllable infinite video generation emerges from autoregressive self-rollout},
  author={Yesiltepe, Hidir and Meral, Tuna Han Salih and Akan, Adil Kaan and Oktay, Kaan and Yanardag, Pinar},
  journal={arXiv preprint arXiv:2511.20649},
  year={2025}
}

@article{lu2025reward,
  title={Reward forcing: Efficient streaming video generation with rewarded distribution matching distillation},
  author={Lu, Yunhong and Zeng, Yanhong and Li, Haobo and Ouyang, Hao and Wang, Qiuyu and Cheng, Ka Leong and Zhu, Jiapeng and Cao, Hengyuan and Zhang, Zhipeng and Zhu, Xing and others},
  journal={arXiv preprint arXiv:2512.04678},
  year={2025}
}

@article{liu2025rolling,
  title={Rolling forcing: Autoregressive long video diffusion in real time},
  author={Liu, Kunhao and Hu, Wenbo and Xu, Jiale and Shan, Ying and Lu, Shijian},
  journal={arXiv preprint arXiv:2509.25161},
  year={2025}
}

@article{huang2025self,
  title={Self forcing: Bridging the train-test gap in autoregressive video diffusion},
  author={Huang, Xun and Li, Zhengqi and He, Guande and Zhou, Mingyuan and Shechtman, Eli},
  journal={arXiv preprint arXiv:2506.08009},
  year={2025}
}

@article{zhu2026causal,
  title={Causal Forcing: Autoregressive Diffusion Distillation Done Right for High-Quality Real-Time Interactive Video Generation},
  author={Zhu, Hongzhou and Zhao, Min and He, Guande and Su, Hang and Li, Chongxuan and Zhu, Jun},
  journal={arXiv preprint arXiv:2602.02214},
  year={2026}
}

@article{huang2025live,
  title={Live avatar: Streaming real-time audio-driven avatar generation with infinite length},
  author={Huang, Yubo and Guo, Hailong and Wu, Fangtai and Zhang, Shifeng and Huang, Shijie and Gan, Qijun and Liu, Lin and Zhao, Sirui and Chen, Enhong and Liu, Jiaming and others},
  journal={arXiv preprint arXiv:2512.04677},
  year={2025}
}

@article{zhuang2025flashvsr,
  title={Flashvsr: Towards real-time diffusion-based streaming video super-resolution},
  author={Zhuang, Junhao and Guo, Shi and Cai, Xin and Li, Xiaohui and Liu, Yihao and Yuan, Chun and Xue, Tianfan},
  journal={arXiv preprint arXiv:2510.12747},
  year={2025}
}

@article{shin2025motionstream,
  title={Motionstream: Real-time video generation with interactive motion controls},
  author={Shin, Joonghyuk and Li, Zhengqi and Zhang, Richard and Zhu, Jun-Yan and Park, Jaesik and Shechtman, Eli and Huang, Xun},
  journal={arXiv preprint arXiv:2511.01266},
  year={2025}
}

@article{yang2025longlive,
  title={Longlive: Real-time interactive long video generation},
  author={Yang, Shuai and Huang, Wei and Chu, Ruihang and Xiao, Yicheng and Zhao, Yuyang and Wang, Xianbang and Li, Muyang and Xie, Enze and Chen, Yingcong and Lu, Yao and others},
  journal={arXiv preprint arXiv:2509.22622},
  year={2025}
}

@inproceedings{li2023photomaker,
  title={PhotoMaker: Customizing Realistic Human Photos via Stacked ID Embedding},
  author={Li, Zhen and Cao, Mingdeng and Wang, Xintao and Qi, Zhongang and Cheng, Ming-Ming and Shan, Ying},
  booktitle=CVPR,
  year={2024}
}

@inproceedings{vace,
    title = {VACE: All-in-One Video Creation and Editing},
    author = {Jiang, Zeyinzi and Han, Zhen and Mao, Chaojie and Zhang, Jingfeng and Pan, Yulin and Liu, Yu},
    booktitle=ICCV,
    year={2025}
}

@inproceedings{xue2025stand,
  title={Stand-in: A lightweight and plug-and-play identity control for video generation},
  author={Xue, Bowen and Duan, Zheng-Peng and Yan, Qixin and Wang, Wenjing and Liu, Hao and Guo, Chun-Le and Li, Chongyi and Li, Chen and Lyu, Jing},
  booktitle=CVPR,
  year={2026}
}

@article{liu2025phantom,
  title={Phantom: Subject-consistent video generation via cross-modal alignment},
  author={Liu, Lijie and Ma, Tianxiang and Li, Bingchuan and Chen, Zhuowei and Liu, Jiawei and Li, Gen and Zhou, Siyu and He, Qian and Wu, Xinglong},
  journal={arXiv preprint arXiv:2502.11079},
  year={2025}
}

@article{li2025bindweave,
  title={BindWeave: Subject-Consistent Video Generation via Cross-Modal Integration},
  author={Li, Zhaoyang and Qian, Dongjun and Su, Kai and Diao, Qishuai and Xia, Xiangyang and Liu, Chang and Yang, Wenfei and Zhang, Tianzhu and Yuan, Zehuan},
  journal={arXiv preprint arXiv:2510.00438},
  year={2025}
}

@article{deng2025magref,
  title={MAGREF: Masked Guidance for Any-Reference Video Generation with Subject Disentanglement},
  author={Deng, Yufan and Yin, Yuanyang and Guo, Xun and Wang, Yizhi and Fang, Jacob Zhiyuan and Yuan, Shenghai and Yang, Yiding and Wang, Angtian and Liu, Bo and Huang, Haibin and others},
  journal={arXiv preprint arXiv:2505.23742},
  year={2025}
}

@article{fei2025skyreels,
  title={SkyReels-A2: Compose Anything in Video Diffusion Transformers},
  author={Fei, Zhengcong and Li, Debang and Qiu, Di and Wang, Jiahua and Dou, Yikun and Wang, Rui and Xu, Jingtao and Fan, Mingyuan and Chen, Guibin and Li, Yang and others},
  journal={arXiv preprint arXiv:2504.02436},
  year={2025}
}

@article{zhang2025kaleido,
  title={Kaleido: Open-Sourced Multi-Subject Reference Video Generation Model},
  author={Zhang, Zhenxing and Teng, Jiayan and Yang, Zhuoyi and Cao, Tiankun and Wang, Cheng and Gu, Xiaotao and Tang, Jie and Guo, Dan and Wang, Meng},
  journal={arXiv preprint arXiv:2510.18573},
  year={2025}
}

@inproceedings{ho2020denoising,
  title={Denoising diffusion probabilistic models},
  author={Ho, Jonathan and Jain, Ajay and Abbeel, Pieter},
  booktitle=NeurIPS,
  year={2020}
}

@article{song2025lightmotion,
  title={LightMotion: A Light and Tuning-free Method for Simulating Camera Motion in Video Generation},
  author={Song, Quanjian and Lin, Zhihang and Zeng, Zhanpeng and Zhang, Ziyue and Cao, Liujuan and Ji, Rongrong},
  journal={arXiv preprint arXiv:2503.06508},
  year={2025}
}

@article{zhang2025objectadd,
  title={Objectadd: adding objects into image via a training-free diffusion modification fashion},
  author={Zhang, Ziyue and Lin, Mingbao and Song, Quanjian and Zhang, Yuxin and Ji, Rongrong},
  journal=PR,
  year={2025},
}

@article{lari2022artifical,
  title={Artifical intelligence in E-commerce: Applications, implications and challenges},
  author={Lari, Halima Afroz and Vaishnava, Kuhu and Manu, KS},
  journal={Asian Journal of Management},
  year={2022},
}

@article{song2025worldwander,
  title={WorldWander: Bridging Egocentric and Exocentric Worlds in Video Generation},
  author={Song, Quanjian and Song, Yiren and Peng, Kelly and Gao, Yuan and Shou, Mike Zheng},
  journal={arXiv preprint arXiv:2511.22098},
  year={2025}
}

@inproceedings{fu2024drivegenvlm,
  title={Drivegenvlm: Real-world video generation for vision language model based autonomous driving},
  author={Fu, Yongjie and Jain, Anmol and Chen, Xu and Mo, Zhaobin and Di, Xuan},
  booktitle={IEEE International automated vehicle validation conference},
  year={2024}
}

@inproceedings{li2025realcam,
  title={Realcam-i2v: Real-world image-to-video generation with interactive complex camera control},
  author={Li, Teng and Zheng, Guangcong and Jiang, Rui and Zhan, Shuigen and Wu, Tao and Lu, Yehao and Lin, Yining and Deng, Chuanyun and Xiong, Yepan and Chen, Min and others},
  booktitle=ICCV,
  year={2025}
}

@inproceedings{hu2022lora,
  title={Lora: Low-rank adaptation of large language models.},
  author={Hu, Edward J and Shen, Yelong and Wallis, Phillip and Allen-Zhu, Zeyuan and Li, Yuanzhi and Wang, Shean and Wang, Liang and Chen, Weizhu and others},
  booktitle=ICLR,
  year={2022}
}

@inproceedings{wu2022fast,
  title={Fast-vqa: Efficient end-to-end video quality assessment with fragment sampling},
  author={Wu, Haoning and Chen, Chaofeng and Hou, Jingwen and Liao, Liang and Wang, Annan and Sun, Wenxiu and Yan, Qiong and Lin, Weisi},
  booktitle=ECCV,
  year={2022},
}

@article{li2025unimatch,
  title={Unimatch: Universal matching from atom to task for few-shot drug discovery},
  author={Li, Ruifeng and Li, Mingqian and Liu, Wei and Zhou, Yuhua and Zhou, Xiangxin and Yao, Yuan and Zhang, Qiang and Chen, Hongyang},
  journal={arXiv preprint arXiv:2502.12453},
  year={2025}
}

@article{wu2023q,
  title={Q-align: Teaching lmms for visual scoring via discrete text-defined levels},
  author={Wu, Haoning and Zhang, Zicheng and Zhang, Weixia and Chen, Chaofeng and Liao, Liang and Li, Chunyi and Gao, Yixuan and Wang, Annan and Zhang, Erli and Sun, Wenxiu and others},
  journal={arXiv preprint arXiv:2312.17090},
  year={2023}
}

@inproceedings{ke2021musiq,
  title={Musiq: Multi-scale image quality transformer},
  author={Ke, Junjie and Wang, Qifei and Wang, Yilin and Milanfar, Peyman and Yang, Feng},
  booktitle=ICCV,
  year={2021}
}

@article{wang2024qwen2,
  title={Qwen2-vl: Enhancing vision-language model's perception of the world at any resolution},
  author={Wang, Peng and Bai, Shuai and Tan, Sinan and Wang, Shijie and Fan, Zhihao and Bai, Jinze and Chen, Keqin and Liu, Xuejing and Wang, Jialin and Ge, Wenbin and others},
  journal={arXiv preprint arXiv:2409.12191},
  year={2024}
}

@inproceedings{deng2019arcface,
  title={Arcface: Additive angular margin loss for deep face recognition},
  author={Deng, Jiankang and Guo, Jia and Xue, Niannan and Zafeiriou, Stefanos},
  booktitle=CVPR,
  year={2019}
}

@inproceedings{huang2024vbench,
  title={Vbench: Comprehensive benchmark suite for video generative models},
  author={Huang, Ziqi and He, Yinan and Yu, Jiashuo and Zhang, Fan and Si, Chenyang and Jiang, Yuming and Zhang, Yuanhan and Wu, Tianxing and Jin, Qingyang and Chanpaisit, Nattapol and others},
  booktitle=CVPR,
  year={2024}
}

@article{yuan2025opens2v,
  title={Opens2v-nexus: A detailed benchmark and million-scale dataset for subject-to-video generation},
  author={Yuan, Shenghai and He, Xianyi and Deng, Yufan and Ye, Yang and Huang, Jinfa and Lin, Bin and Luo, Jiebo and Yuan, Li},
  journal={arXiv preprint arXiv:2505.20292},
  year={2025}
}

@article{he2024id,
  title={Id-animator: Zero-shot identity-preserving human video generation},
  author={He, Xuanhua and Liu, Quande and Qian, Shengju and Wang, Xin and Hu, Tao and Cao, Ke and Yan, Keyu and Zhang, Jie},
  journal={arXiv preprint arXiv:2404.15275},
  year={2024}
}

@inproceedings{yuan2025identity,
  title={Identity-preserving text-to-video generation by frequency decomposition},
  author={Yuan, Shenghai and Huang, Jinfa and He, Xianyi and Ge, Yunyang and Shi, Yujun and Chen, Liuhan and Luo, Jiebo and Yuan, Li},
  booktitle=CVPR,
  year={2025}
}

@inproceedings{chen2024disenstudio,
  title={Disenstudio: Customized multi-subject text-to-video generation with disentangled spatial control},
  author={Chen, Hong and Wang, Xin and Zhang, Yipeng and Zhou, Yuwei and Zhang, Zeyang and Tang, Siao and Zhu, Wenwu},
  booktitle=ACMMM,
  year={2024}
}

@article{wang2026customvideo,
  title={Customvideo: Customizing text-to-video generation with multiple subjects},
  author={Wang, Zhao and Li, Aoxue and Zhu, Lingting and Guo, Yong and Dou, Qi and Li, Zhenguo},
  journal=TMM,
  year={2026},
}

@inproceedings{wang2024motionctrl,
  title={Motionctrl: A unified and flexible motion controller for video generation},
  author={Wang, Zhouxia and Yuan, Ziyang and Wang, Xintao and Li, Yaowei and Chen, Tianshui and Xia, Menghan and Luo, Ping and Shan, Ying},
  booktitle=SIGGRAPH,
  year={2024}
}

@article{song2025univst,
  title={Univst: A unified framework for training-free localized video style transfer},
  author={Song, Quanjian and Lin, Mingbao and Zhan, Wengyi and Yan, Shuicheng and Cao, Liujuan and Ji, Rongrong},
  journal=PAMI,
  year={2025},
}

@inproceedings{peebles2023scalable,
  title={Scalable diffusion models with transformers},
  author={Peebles, William and Xie, Saining},
  booktitle=ICCV,
  year={2023}
}

@inproceedings{bao2023all,
  title={All are worth words: A vit backbone for diffusion models},
  author={Bao, Fan and Nie, Shen and Xue, Kaiwen and Cao, Yue and Li, Chongxuan and Su, Hang and Zhu, Jun},
  booktitle=CVPR,
  year={2023}
}

@article{gao2024ca2,
  title={Ca2-vdm: Efficient autoregressive video diffusion model with causal generation and cache sharing},
  author={Gao, Kaifeng and Shi, Jiaxin and Zhang, Hanwang and Wang, Chunping and Xiao, Jun and Chen, Long},
  journal={arXiv preprint arXiv:2411.16375},
  year={2024}
}

@article{hu2024acdit,
  title={Acdit: Interpolating autoregressive conditional modeling and diffusion transformer},
  author={Hu, Jinyi and Hu, Shengding and Song, Yuxuan and Huang, Yufei and Wang, Mingxuan and Zhou, Hao and Liu, Zhiyuan and Ma, Wei-Ying and Sun, Maosong},
  journal={arXiv preprint arXiv:2412.07720},
  year={2024}
}

@article{zhang2025test,
  title={Test-time training done right},
  author={Zhang, Tianyuan and Bi, Sai and Hong, Yicong and Zhang, Kai and Luan, Fujun and Yang, Songlin and Sunkavalli, Kalyan and Freeman, William T and Tan, Hao},
  journal={arXiv preprint arXiv:2505.23884},
  year={2025}
}
